\pdfoutput=1
\PassOptionsToPackage{table}{xcolor}
\documentclass[11pt]{article}

\usepackage[final]{acl}
\usepackage[table]{xcolor}
\usepackage{array,tabularx,booktabs}

\usepackage{xcolor}
\usepackage{times}
\usepackage{latexsym}
\usepackage{amsmath}
\usepackage{mathtools} 
\usepackage{booktabs}
\usepackage[table]{xcolor}
\usepackage{multirow}
\usepackage{array}
\usepackage{float}
\usepackage{makecell}
\usepackage{caption}
\usepackage[most]{tcolorbox}
\usepackage{xcolor}
\usepackage{booktabs}
\usepackage{xcolor}
\usepackage{pifont}
\usepackage{siunitx}
\usepackage{amsmath} 
\usepackage{xcolor}
\usepackage{siunitx}
\usepackage{threeparttable}
\usepackage{booktabs,multirow,threeparttable}
\usepackage{siunitx}
\usepackage[table]{xcolor}
\usepackage{colortbl}
\usepackage{nicematrix}

\usepackage[table]{xcolor}

\usepackage{amsmath,amssymb,amsthm}
\DeclareMathOperator*{\argmax}{arg\,max}
\definecolor{forestgreen}{RGB}{34, 139, 34}
\definecolor{brickred}{RGB}{178, 34, 34}
\definecolor{darkorange}{RGB}{255, 140, 0}
\definecolor{oursrow}{HTML}{D9ECFF}

\newcommand{\doublerule}{%
  \specialrule{0.4pt}{0pt}{1.2pt}%
  \specialrule{0.4pt}{0pt}{0pt}%
}

\usepackage[ruled,vlined]{algorithm2e}

\usepackage[T1]{fontenc}

\usepackage[utf8]{inputenc}

\usepackage{microtype}

\usepackage{inconsolata}

\usepackage{graphicx}

\title{\raisebox{-0.2\height}{\includegraphics[height=1.75em]{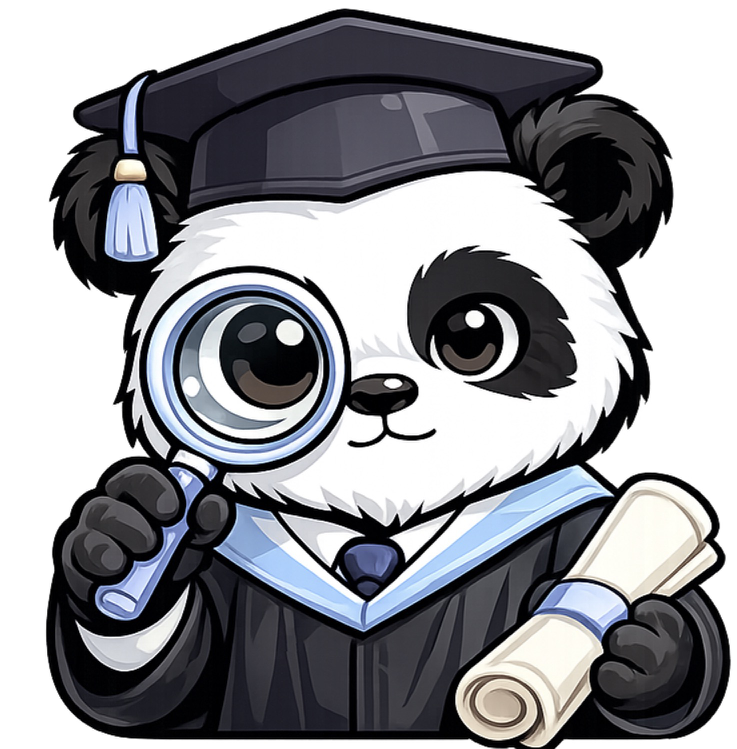}}\hspace{0.2em}BizSage: A Self-Evolving Multi-Agent Framework for Business Research with Efficient Knowledge Retrieval}

\author{
 \textbf{Yuhe Wu}\thanks{Equal contribution.},
 \textbf{Guangyu Wang}\footnotemark[1],
 \textbf{Jiaxin Liu}\footnotemark[1],
 \textbf{Guang Zhang}\thanks{Corresponding author.}
\\[0.55em]
 The Hong Kong University of Science and Technology (Guangzhou), Guangzhou, China
\\[0.45em]
 \raisebox{-0.4em}{\includegraphics[height=1.45em]{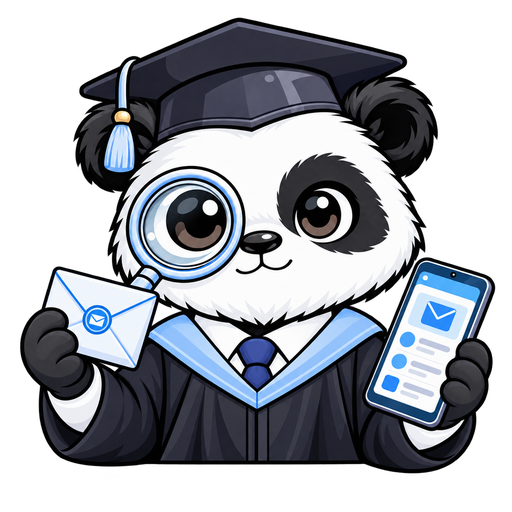}}:\hspace{0.35em}\textcolor{darkblue}{\{ywu724@connect.,\,guangzhang@\}hkust-gz.edu.cn}
\\[0.3em]
 \raisebox{-0.4em}{\includegraphics[height=1.45em]{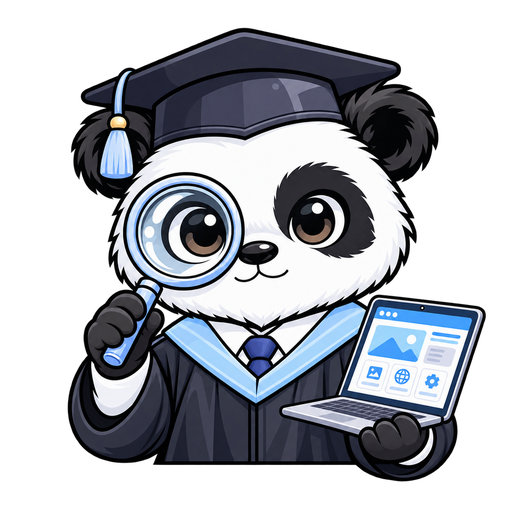}}:\hspace{0.35em}\href{https://bizsage.site}{\textcolor{darkblue}{https://bizsage.site}}
}

\begin{document}

\maketitle

\begin{abstract}
While multi-agent systems based on large language models (LLMs) have shown promise in automating the progressive workflow of academic research, extending them to economics and business research, where specialized domain knowledge spans neighboring disciplines yet remains difficult to access in a structured way, presents two challenges. First, existing methods mostly retrieve at the paper level, yet the evidence needed for research tasks is often distributed across different sections, creating a granularity mismatch that hinders retrieval coverage and precision. Second, these fields demand strict empirical rigor, yet current systems provide limited mechanisms for learning from evaluation feedback. We present \textbf{BizSage}, a multi-agent framework combining corpus-level fine-grained retrieval with quality-driven self-evolution. We build a Lateral Knowledge Graph (LKG) by merging section-level knowledge graphs and apply Personalized PageRank (PPR) to surface semantically relevant and structurally important sections. Seven specialized agents collaborate under a Meta-Review self-evolution mechanism that distills failure modes from evaluation traces into reusable strategies. On a benchmark spanning four domains and three tasks, BizSage ranks first on the majority of metrics, achieves pairwise win-rates above 60\% against six baselines, and produces zero hallucinated citations. We hope BizSage paves the way for reliable research assistance in economics, business, and the broader social sciences.
\end{abstract}

\section{Introduction}
\label{sec:intro}

Academic research typically follows progressive convergence.
Researchers first survey the literature to understand the state of a field,
then identify open problems and form research ideas, and refine
those ideas into experimental plans through iterative
improvement~\citep{snyder2019literature, fernandez2019critically, duncan2023research}.
This process requires locating relevant work in growing literature,
identifying gaps, and integrating methodological knowledge across disciplines.
As the volume of academic publications grows, the cognitive cost of literature
screening, method comparison, and research design continues to rise, making
automated research assistance a common need in interdisciplinary
work~\citep{harari2020literature,shahrzadi2024causes,finocchi2025large}.
This challenge is particularly pronounced in economics and business research,
where relevant knowledge is often distributed across neighboring disciplines,
raising the bar for accurate literature retrieval and method
selection~\citep{kapeller2017citation, gusenbauer2020academic, gusenbauer2021every}.

Multi-agent systems driven by large language models (LLMs) have
shown potential for automated scientific
discovery~\citep{boiko2023autonomous, ghafarollahi2025sciagents}.
Current research proceeds along two main directions.
The first focuses on idea generation, using multi-agent collaboration to
propose, review, and refine
hypotheses~\citep{zhou2024hypothesis, qi2024large, baek2025researchagent};
representative frameworks include
VirSci~\citep{su2025many} and Co-Scientist~\citep{gottweis2025towards}.
The second pursues end-to-end systems that automate the full research
workflow~\citep{weng2024cycleresearcher, ghareeb2025robin, mitchener2025kosmos, novikov2025alphaevolve},
such as AI Scientist~\citep{lu2024ai} and its successor~\citep{yamada2025ai},
AI-Researcher~\citep{tang2025ai}, and InternAgent~\citep{team2025internagent}.
Building on these frameworks, recent work addresses the evolution of system
capabilities and knowledge
retention~\citep{weidener2026rethinking, lin2026scider, bicker2026aster, feng2026internagent};
EvoScientist~\citep{lyu2026evoscientist} uses persistent memory to enable
cross-task capability accumulation.
These advances show that LLM-based agents are nearing the capability
for autonomous research.

\begin{figure}[t]
  \centering
  \includegraphics[width=0.45\textwidth]{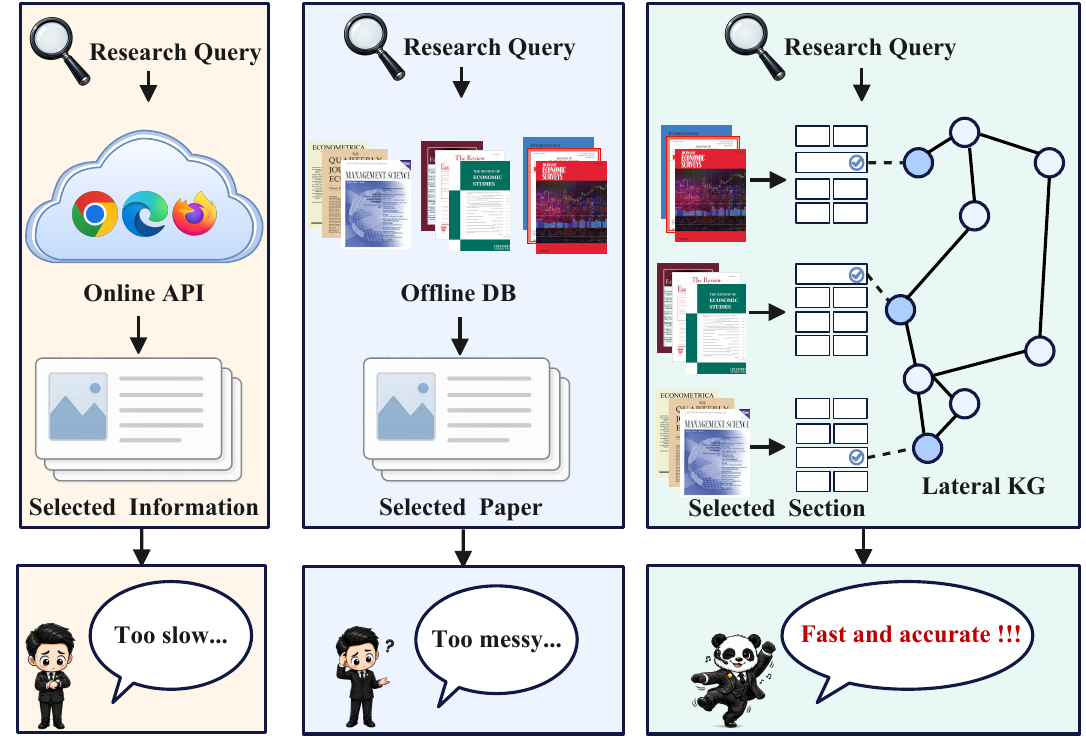}
  \caption{Comparison of three retrieval methods. \textbf{Left}: online API search returns selected information but is slow. \textbf{Middle}: offline database retrieves whole papers but includes irrelevant content. \textbf{Right}: our approach builds a Lateral Knowledge Graph and retrieves selected sections, achieving both speed and accuracy.}
  \label{fig:teaser}
\end{figure}

Although these systems show promise in automating research tasks,
extending them to economics and business research presents two
challenges, as illustrated in Figure~\ref{fig:teaser}.
First, existing methods mostly retrieve at the paper level, yet the
evidence needed for literature review and research design often
resides in scattered sections within a single paper.
In large-scale interdisciplinary corpora, this granularity mismatch makes
it difficult to balance retrieval coverage, precision, and efficiency.
Second, economics and business research demands strict empirical rigor,
requiring systems to produce quality research plans and
adapt to domain-specific quality patterns.
Accordingly, we study the following research question:

\begin{tcolorbox}[
  enhanced,
  colback=blue!3,
  colframe=blue!35,
  boxrule=0.5pt,
  arc=2pt,
  left=8pt, right=8pt, top=6pt, bottom=6pt,
]
\textcolor{blue!50}{\ding{118}}
\textit{How can we build a multi-agent system for economics and business
research that achieves efficient and accurate fine-grained knowledge retrieval
while continuously adapting to domain-specific quality standards?}
\end{tcolorbox}

To address this question, we propose \textbf{BizSage}, a self-evolving
multi-agent framework for economics and business research assistance.
The framework pre-builds a corpus-level Lateral Knowledge Graph (LKG)
from over 130{,}000 source papers spanning economics, finance,
operations management, and statistics, and applies Personalized PageRank (PPR)
to retrieve sections that are both relevant and structurally important.
Multiple specialized agents then collaborate on survey, idea
formulation, and research plan as progressive tasks, guided by a
Meta-Review self-evolution mechanism that distills reusable quality
strategies from historical evaluations.

Our contributions are summarized as follows:
\begin{itemize}
  \item We propose BizSage, to our knowledge the first multi-agent framework with built-in self-evolution for economics and business research, and construct a 300-query benchmark spanning four domains. BizSage achieves leading scores on the majority of metrics across three backbone LLMs.
  \item We introduce an efficient fine-grained retrieval mechanism that unifies section-level knowledge graphs into a corpus-level graph and applies PPR to capture cross-document evidence that paper-level retrieval misses.
  \item We propose a Meta-Review self-evolution mechanism that converts accumulated evaluation feedback into actionable quality strategies, enabling progressive adaptation to domain-specific quality standards.
  \item We extend citation hallucination evaluation from standalone LLMs to multi-agent research systems, identify domain drift as a new failure mode, and demonstrate zero hallucinated citations.
\end{itemize}

\section{Business Automated Research Task}

We highlight that relevant literature retrieval is critical to these tasks, as language models lack sufficient domain-specific knowledge to conduct rigorous research independently. Accordingly, we formalize these tasks as a two-stage process: \textbf{Stage 1}: literature retrieval, which gathers relevant domain evidence for a given research request; and \textbf{Stage 2}: grounded response generation, which produces a response supported by the retrieved evidence.

\subsection{Problem Definition}
Let $\mathcal{Q} = \{q_1, q_2, \ldots, q_m\}$ be a set of heterogeneous research tasks (e.g., survey, idea formulation, research planning), and let $\mathcal{C}$ be a domain-specific corpus. For each $q_i \in \mathcal{Q}$, the system first retrieves a task-specific literature collection $\mathcal{D}_i = \{d_1, d_2, \ldots, d_n\} \subseteq \mathcal{C}$, where each $d_j$ is a document pertinent to $q_i$. The objective is then to generate the optimal response $y_i^*$:
\begin{align}
    \mathcal{D}_i &= \mathrm{Retrieve}(q_i,\, \mathcal{C}) \notag \\
    y_i^* &= \argmax_{y}\, P(y \mid q_i,\, \mathcal{D}_i) \label{eq:problem}
\end{align}

\section{Method}
\label{sec:method}
\begin{figure*}[t]
  \centering
  \includegraphics[width=0.95\textwidth]{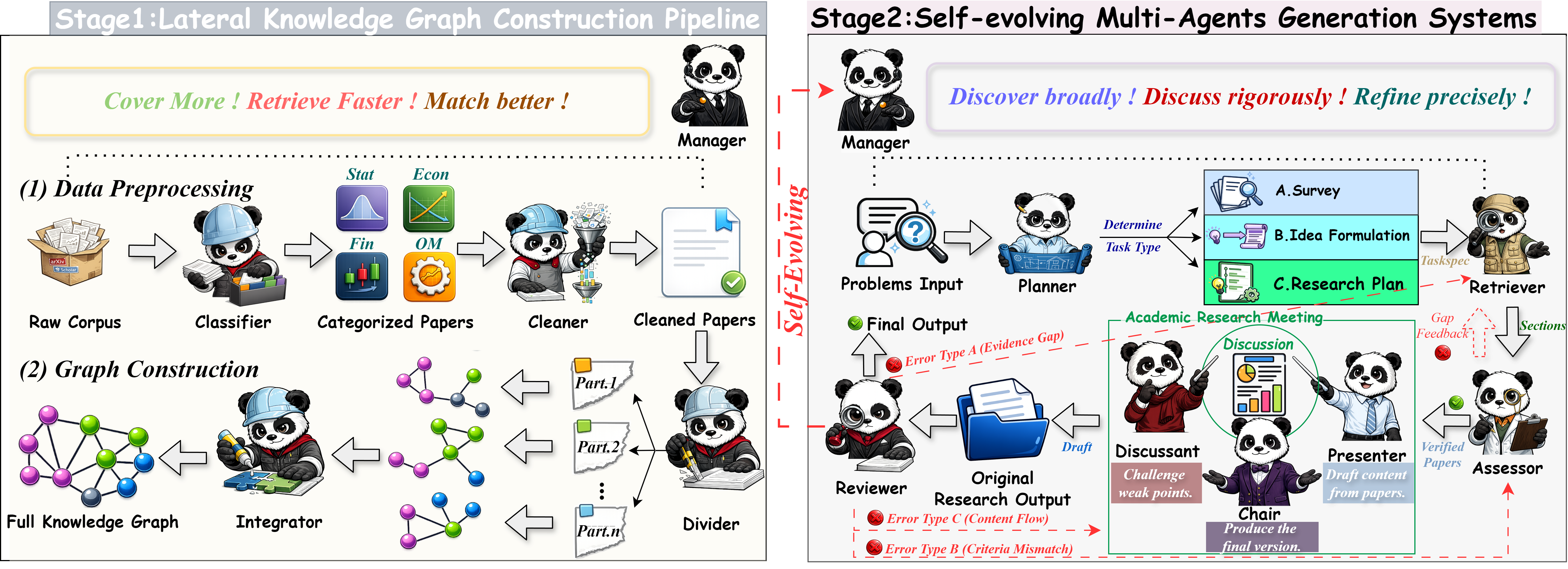}
  \caption{Overview of the proposed framework. The left part illustrates \textbf{Stage 1} (literature retrieval) and the right part illustrates \textbf{Stage 2} (grounded response generation).}
  \label{fig:framework}
\end{figure*}

In this section, we detail the proposed framework, as illustrated in Figure~\ref{fig:framework}.

\subsection{Stage 1: Literature Retrieval via Knowledge Graph}

Retrieving the most relevant content from a large corpus is non-trivial~\citep{gusenbauer2020academic}.
A naive similarity search tends to surface many semantically similar but informationally redundant sections, missing the truly important ones. To address this, we propose a Lateral Knowledge Graph (LKG), a single unified graph built across all documents in the corpus. LKG allows concept nodes from different papers to be connected, enabling cross-document knowledge flow during retrieval. We then apply Personalized PageRank (PPR)~\citep{10.1109/TKDE.2024.3376000} on the LKG to identify sections that are both relevant and structurally important.

\paragraph{Section-Level Knowledge Graph Extraction.}
For each document $d_j \in \mathcal{C}$, we segment it into sections and prompt a LLM to extract a local knowledge graph $G_j = (V_j, E_j)$ from each section, where nodes $V_j$ represent key concepts and edges $E_j$ represent semantic relations between them~\citep{lairgi2024itext2kg,mo2025kggen}.

\paragraph{Lateral Knowledge Graph Construction.}
Duplicate or semantically equivalent nodes are inevitable in section-level graph construction, yet prior methods often overlook this issue, leading to fragmented graphs and limited effectiveness. We therefore introduce an additional merging step to align these nodes across documents and construct a unified corpus-level lateral knowledge graph $G = (V, E)$:
\begin{equation}
    G = \bigcup_{j} G_j
    \label{eq:kg_merge}
\end{equation}
This design consolidates concepts shared across multiple papers into unified nodes, making cross-paper connections explicit and allowing PPR to propagate importance signals across the entire corpus rather than within individual documents.

\paragraph{Personalized PageRank Retrieval.}
Given a query $q_i$, we initialize a personalization vector $\mathbf{p}$ over nodes in $G$ based on the semantic similarity between $q_i$ and each concept node. We then run PPR on $G$ to obtain an importance score $\pi(v)$ for each node $v$:
\begin{equation}
    \boldsymbol{\pi} = \alpha \mathbf{p} + (1-\alpha) \mathbf{A}^\top \boldsymbol{\pi}
    \label{eq:ppr}
\end{equation}
where $\mathbf{A}$ is the column-normalized adjacency matrix of $G$ and $\alpha$ is the teleportation probability. Sections whose constituent nodes accumulate high PPR scores are ranked as the most relevant and important, effectively distinguishing informative sections from merely similar but redundant ones. The top-$n$ sections are collected to form the retrieved literature $\mathcal{D}_i$.

\subsection{Stage 2: Self-Evolving Multi-Agent Generation System}

Stage~2 organizes Equation~\eqref{eq:problem} into seven agents coordinated by a Manager.
The Manager maintains a \emph{quality strategy memory} $M_Q$ indexed by agent and weakness type; before each task, every agent retrieves its strategies $K_a = \mathcal{R}_Q(M_Q, a)$.

\paragraph{Planner Agent.}
The Planner maps each input $q_i$ to a task specification $(\tau_i, z_i)$, where $\tau_i$ identifies the task type and $z_i$ is the research intent.

\paragraph{Retriever Agent.}
The Retriever applies PPR on the LKG from Stage~1 to collect $n$ candidate sections $\mathcal{D}_i$.
When the Assessor judges evidence insufficient, the Retriever receives new keywords and retrieves additional sections, skipping seen papers.

\paragraph{Assessor Agent.}
The Assessor labels each $d_j \in \mathcal{D}_i$ as useful $\mathcal{D}_i^+$, uncertain $\mathcal{D}_i^{\pm}$, or noise $\mathcal{D}_i^-$; evidence is sufficient when $|\mathcal{D}_i^+| \geq \theta$.
It then ranks $\mathcal{D}_i^+$ by relevance and diversity and selects the top $M_{\tau}$ sections, where $M_{\tau}$ is set per task type:
\begin{equation}
    \mathcal{D}_i^{\mathrm{write}} = \mathrm{Top}_{M_{\tau}}\!\bigl(\mathcal{D}_i^+\bigr).
    \label{eq:rank}
\end{equation}

\paragraph{Academic Research Meeting.}
The writing stage proceeds in three steps. The \textbf{Presenter} generates a draft $\hat{y}_i^{(0)}$ from $(\tau_i, z_i, \mathcal{D}_i^{\mathrm{write}})$ with a task-specific template. The \textbf{Discussant} examines it on multiple quality dimensions and raises issues $\mathcal{R}_i$. The \textbf{Chair} integrates the draft and critique into a revised output:
\begin{equation}
    \hat{y}_i = g\!\bigl(\hat{y}_i^{(0)},\, \mathcal{R}_i,\, \mathcal{D}_i^{\mathrm{write}}\bigr).
    \label{eq:chair}
\end{equation}

\paragraph{Reviewer Agent.}
The Reviewer scores $\hat{y}_i$ on multiple dimensions, yielding $\mathbf{s}_i \in \{1,\ldots,5\}^4$ with total $S_i = \sum_{k} s_{i,k}$.
The output is accepted when $S_i \geq S_{\mathrm{pass}}$ and $\min_k s_{i,k} \geq s_{\min}$; otherwise it identifies the responsible agent $a_i^*$ and generates feedback $f_{a_i^*}$.

\paragraph{Self-Evolving Manager.}
The Manager enables agents to evolve at two granularities. At the task level, it corrects the current output by routing Reviewer feedback to the diagnosed agent and re-running from that point for up to $T$ rounds, where $\Phi$ denotes one pipeline pass:
\begin{equation}
    \hat{y}_i^{(t+1)} = \Phi(q_i \mid f_{a^*}^{(t)}),
    \quad t = 1, \ldots, T.
    \label{eq:iterate}
\end{equation}
At the system level, after $m$ tasks the Manager analyzes traces $\{H_1, \ldots, H_m\}$ to identify agents that are repeatedly blamed and the failure patterns behind them.
For each pattern, it generates candidate strategies $\{c_1, \ldots, c_K\}$ and selects the one that most improves the worst historical case:
\begin{equation}
    c_a^* = \argmax_{k}\, \nu(c_k,\, H_{\mathrm{worst}}),
    \label{eq:replay}
\end{equation}
where $\nu$ measures quality gain. Validated strategies are committed to $M_Q$:
\begin{equation}
    M_Q \leftarrow M_Q \cup \{(a,\, c_a^*)\}.
    \label{eq:update_mq}
\end{equation}
This distills recurring failures into general strategies that change agent behavior, rather than storing individual past outputs. The prompt templates for all agents are provided in Appendix~\ref{app:prompts}.

\section{Experimental Setup}
\label{sec:experiments}
To evaluate \textbf{BizSage} and understand what drives its performance on domain-specific research tasks, we investigate three research questions (RQs):

\begin{itemize}
\item \textbf{RQ1:} How well does BizSage meet the quality standards of business and economics scholarship compared to existing methods?
\item \textbf{RQ2:} How does self-evolution reshape the quality profile across iterations, and does backbone capacity affect the improvement ceiling?
\item \textbf{RQ3:} Which aspects of output quality depend on structured retrieval, and which require multi-agent collaboration?
\end{itemize}

\subsection{Data Source}

\paragraph{Knowledge Graph.}
For the knowledge graph construction in Section~\ref{sec:method}, we employ GPT-4o to extract entities and relations from each paper section.
Following prior work~\cite{zhu2024llms}, we treat knowledge graph extraction as a fundamental information extraction task where GPT-4o is sufficient.
After merging all section-level graphs across the corpus, the resulting LKG contains 3{,}799{,}002 concept nodes and 9{,}722{,}116 edges.
Details of our dataset are provided in Appendix~\ref{app:journals}.

\paragraph{Evaluation Dataset.}
We hold out 4{,}000 papers from the source corpus before LKG construction and derive 600 research queries from them, preventing information leakage between the retrieval index and evaluation targets.
The three task types are \textbf{\emph{survey}}, which asks for a structured review of a research area; \textbf{\emph{idea formulation}}, which asks for a novel methodological proposal addressing a specific problem; and \textbf{\emph{research plan}}, which asks for a concrete experimental design with hypotheses, data, and validation strategy.
Each type yields 200 queries across four domains, from which 100 are randomly sampled as the development set for self-evolution and the remaining 100 form the evaluation benchmark. The evaluation set remains unseen throughout the evolution process. Details of the construction pipeline appear in Appendix~\ref{app:benchmark}.

\subsection{Baselines}
We compare against two groups of baselines.
Full-pipeline systems that support all three task types include AI-Scientist-v2~\citep{yamada2025ai}, EvoScientist~\citep{lyu2026evoscientist}, InternAgent-1.5~\citep{feng2026internagent} and ResearchAgent~\citep{baek2025researchagent}.
Task-specific systems serve as additional baselines for their respective task types: Idea2Story~\citep{xu2026idea2story} for idea formulation, and AI Co-Scientist~\citep{gottweis2025towards} for research plan.
Implementation details for each baseline appear in Appendix~\ref{app:baselines}.

\subsection{Evaluation Metrics}
We use GPT-4o as an LLM judge~\citep{liu2023g, yu2025improve} to score each output using a 1--5 scale.
All task types are evaluated on \textbf{Relevance} and \textbf{Grounding}, measuring query alignment and citation reliability, respectively~\citep{agarwal2024litllm, wu2025automated}.
The remaining two dimensions are task-specific: \textbf{Coverage} and \textbf{Synthesis} for survey~\citep{snyder2019literature, yan2025surveyforge}, \textbf{Novelty} and \textbf{Feasibility} for idea formulation~\citep{ baek2025researchagent,lyu2026evoscientist}, and \textbf{Rigor} and \textbf{Feasibility} for research plan~\citep{jing2025development, goel2025training}.
Appendix~\ref{app:eval-metrics} provides the full rubric and Appendix~\ref{app:judge-validation} validates LLM--human agreement.

\section{Experimental Results}

\subsection{Main Results (RQ1)}

\begin{table*}[t]
\centering
\footnotesize
\setlength{\tabcolsep}{3.6pt}
\begin{tabular}{@{}l|cccc|cccc|cccc@{}}
\doublerule
 & \multicolumn{4}{c|}{Survey} & \multicolumn{4}{c|}{Idea Formulation} & \multicolumn{4}{c}{Research Plan} \\
\cmidrule(lr){2-5} \cmidrule(lr){6-9} \cmidrule(lr){10-13}
\multirow{-2}{*}{\textbf{Method}} & Rel. & Grd. & Cov. & Syn. & Rel. & Grd. & Nov. & Fea. & Rel. & Grd. & Rig. & Fea. \\
\midrule
\multicolumn{13}{@{}l}{\textbf{\textit{GPT-5.5}}} \\
AI-Scientist-v2 & \underline{4.68} & \underline{3.57} & \underline{3.97} & \underline{3.94} & 4.68 & \underline{3.29} & \textbf{3.69} & \underline{3.85} & \underline{4.88} & \underline{4.22} & \underline{4.05} & \underline{3.84} \\
EvoScientist & 4.55 & 2.20 & \underline{3.97} & 3.82 & \textbf{4.72} & 2.28 & 2.53 & 3.75 & 4.43 & 1.84 & 2.80 & 2.76 \\
InternAgent-1.5 & 3.52 & 1.97 & 2.54 & 2.84 & 4.19 & 2.16 & 3.04 & 3.36 & 4.22 & 1.56 & 3.87 & 3.03 \\
ResearchAgent & 4.37 & 2.47 & 3.81 & 3.82 & 4.33 & 2.87 & 3.32 & 3.66 & 4.83 & 2.55 & 3.17 & 3.52 \\
\cmidrule{1-13}
Idea2Story$^\dagger$ & -- & -- & -- & -- & 4.66 & 2.56 & 3.14 & 3.69 & -- & -- & -- & -- \\
AI Co-Scientist$^\dagger$ & -- & -- & -- & -- & -- & -- & -- & -- & 3.91 & 1.62 & 2.40 & 2.47 \\
\cmidrule{1-13}
\rowcolor{oursrow} \textbf{BizSage (Ours)} & \textbf{4.77} & \textbf{4.14} & \textbf{4.04} & \textbf{3.99} & \underline{4.71} & \textbf{3.42} & \underline{3.43} & \textbf{3.95} & \textbf{4.89} & \textbf{4.36} & \textbf{4.07} & \textbf{3.95} \\
\midrule
\multicolumn{13}{@{}l}{\textbf{\textit{Gemini-3-Flash}}} \\
AI-Scientist-v2 & 3.73 & \underline{2.48} & 3.00 & \underline{2.85} & 4.09 & 2.25 & 2.85 & 2.72 & \textbf{4.25} & 1.80 & 2.88 & 2.83 \\
EvoScientist & \underline{3.91} & 2.27 & \underline{3.16} & 2.83 & 4.23 & \underline{2.47} & 2.43 & \underline{3.32} & 3.97 & \underline{2.30} & 2.28 & 2.52 \\
InternAgent-1.5 & 2.69 & 1.53 & 2.12 & 1.90 & 3.44 & 1.25 & 2.15 & 2.24 & 3.21 & 1.51 & 2.11 & 2.03 \\
ResearchAgent & 3.67 & 1.82 & 2.75 & 2.53 & 3.65 & 2.19 & 2.76 & 2.74 & 3.99 & 1.96 & \underline{2.89} & \underline{2.84} \\
\cmidrule{1-13}
Idea2Story$^\dagger$ & -- & -- & -- & -- & \textbf{4.44} & 2.03 & \textbf{3.05} & 3.12 & -- & -- & -- & -- \\
AI Co-Scientist$^\dagger$ & -- & -- & -- & -- & -- & -- & -- & -- & 2.58 & 1.36 & 1.23 & 1.48 \\
\cmidrule{1-13}
\rowcolor{oursrow} \textbf{BizSage (Ours)} & \textbf{3.94} & \textbf{2.96} & \textbf{3.41} & \textbf{3.01} & \underline{4.25} & \textbf{2.99} & \underline{2.86} & \textbf{3.78} & \underline{4.02} & \textbf{2.95} & \textbf{2.94} & \textbf{2.86} \\
\midrule
\multicolumn{13}{@{}l}{\textbf{\textit{DeepSeek-V4-Flash}}} \\
AI-Scientist-v2 & 3.93 & 2.61 & 3.06 & 3.03 & 4.37 & 2.62 & 2.91 & \underline{3.31} & \underline{4.46} & 2.06 & 3.42 & 3.28 \\
EvoScientist & \underline{4.51} & \underline{2.76} & \textbf{3.97} & \underline{3.90} & \underline{4.72} & \underline{2.91} & \textbf{3.19} & 3.20 & \textbf{4.87} & \underline{3.67} & \underline{3.89} & \underline{3.68} \\
InternAgent-1.5 & 2.38 & 1.42 & 1.82 & 1.87 & 3.95 & 1.96 & 2.64 & 2.75 & 3.90 & 1.46 & 2.68 & 2.54 \\
ResearchAgent & 3.97 & 2.05 & 3.12 & 3.06 & 3.93 & 2.21 & 2.80 & 3.04 & 4.44 & 2.32 & 3.33 & 3.26 \\
\cmidrule{1-13}
Idea2Story$^\dagger$ & -- & -- & -- & -- & 4.50 & 2.22 & 2.92 & 3.13 & -- & -- & -- & -- \\
AI Co-Scientist$^\dagger$ & -- & -- & -- & -- & -- & -- & -- & -- & 3.59 & 1.46 & 2.09 & 2.23 \\
\cmidrule{1-13}
\rowcolor{oursrow} \textbf{BizSage (Ours)} & \textbf{4.65} & \textbf{2.96} & \underline{3.89} & \textbf{4.08} & \textbf{4.74} & \textbf{3.07} & \underline{3.06} & \textbf{3.79} & \underline{4.46} & \textbf{3.88} & \textbf{3.93} & \textbf{3.70} \\
\doublerule
\end{tabular}
\caption{Main results across three backbone LLMs. \textbf{Bold} and \underline{underline} mark the best and second-best scores in each column within each backbone. $\dagger$ denotes task-specific systems evaluated only on their target task. Metric abbreviations: Rel.\,=\,Relevance, Grd.\,=\,Grounding, Cov.\,=\,Coverage, Syn.\,=\,Synthesis, Nov.\,=\,Novelty, Fea.\,=\,Feasibility, Rig.\,=\,Rigor.}
\label{tab:main-results}
\end{table*}

\begin{figure*}[!t]
\centering
\includegraphics[width=\textwidth]{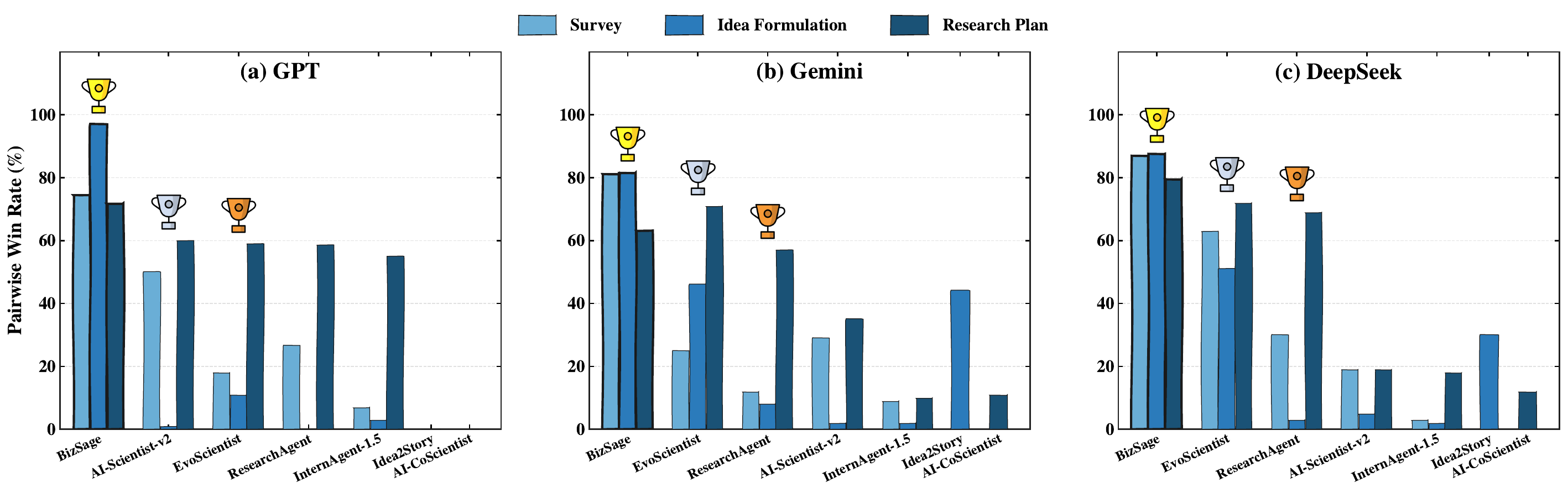}
\caption{Pairwise win-rate (\%) of BizSage against each baseline, judged by GPT-4o. Each sub-figure shows one task type, with bars grouped by backbone.}
\label{fig:pairwise}
\end{figure*}

\paragraph{Quantitative Comparison.}
Table~\ref{tab:main-results} summarizes all methods across three backbones and three task types.
BizSage achieves the highest score on the majority of metrics under every backbone.
On GPT-5.5, it ranks first on \textbf{10 out of 12 metrics}, falling to second only on Relevance and Novelty for Idea Formulation. Among all dimensions, Grounding shows the most pronounced advantage: most baselines score below 3.0 on this dimension, while BizSage leads the second-best method by 0.57 points on Survey.
This gap stems from the cross-document structure of the LKG, where PPR ranking retrieves passages that are both topically relevant and structurally important, thereby reducing citation errors in the generation stage.
The two task-specific baselines target only a single task, yet they fail to consistently outperform BizSage even on their designated task, indicating that a general-purpose framework with strong retrieval and generation can match or exceed specialized systems.

Comparing across backbones, absolute scores for all methods increase with stronger backbone capability, confirming that backbone strength is the key factor determining the upper bound of output quality.
BizSage exhibits a larger relative gain on weaker backbones: its average Grounding lead over the second-best method widens from 0.28 points on the strongest backbone to 0.55 points on the weakest, suggesting that structured retrieval and multi-agent collaboration partially compensate for limited backbone capacity. Representative outputs for each task type are shown in Appendix~\ref{app:examples}.

\paragraph{Pairwise Comparison.}
Independent scoring averages over queries of varying difficulty, which may mask per-query differences.
To control for this, we conduct pairwise evaluation: for each query, GPT-4o directly judges which of two outputs is better, and we report the fraction of queries where BizSage is preferred.
As shown in Figure~\ref{fig:pairwise}, BizSage achieves win-rates above 60\% across most backbone and task combinations, further confirming its broad advantage.

\subsection{Effect of Self-Evolution (RQ2)}

\begin{table*}[t]
\centering
\footnotesize
\setlength{\tabcolsep}{3.6pt}
\renewcommand{\arraystretch}{1.22}
\begin{tabular}{@{}l|cccc|cccc|cccc@{}}
\doublerule
 & \multicolumn{4}{c|}{Survey} & \multicolumn{4}{c|}{Idea Formulation} & \multicolumn{4}{c}{Research Plan} \\
\cmidrule(lr){2-5} \cmidrule(lr){6-9} \cmidrule(lr){10-13}
\multirow{-2}{*}{\textbf{Method}} & Rel. & Grd. & Cov. & Syn. & Rel. & Grd. & Nov. & Fea. & Rel. & Grd. & Rig. & Fea. \\
\midrule
\multicolumn{13}{@{}l}{\textbf{\textit{GPT-5.5}}} \\
BizSage (Iteration 1) & 4.14 & 3.05 & 3.77 & 3.86 & 4.52 & 2.07 & 3.16 & 3.54 & 4.60 & 3.35 & 3.94 & 3.31 \\
BizSage (Iteration 2) & \underline{4.30} & \underline{4.06} & \underline{3.94} & \underline{3.89} & \underline{4.57} & \underline{3.17} & \underline{3.40} & \underline{3.67} & \underline{4.62} & \underline{3.75} & \underline{4.00} & \underline{3.42} \\
\rowcolor{oursrow} BizSage (Iteration 3) & \textbf{4.77} & \textbf{4.14} & \textbf{4.04} & \textbf{3.99} & \textbf{4.71} & \textbf{3.42} & \textbf{3.43} & \textbf{3.95} & \textbf{4.89} & \textbf{4.36} & \textbf{4.07} & \textbf{3.95} \\
\midrule
\multicolumn{13}{@{}l}{\textbf{\textit{Gemini-3-Flash}}} \\
BizSage (Iteration 1) & 3.90 & 2.67 & 3.35 & 2.93 & 4.10 & 2.95 & 2.73 & \underline{3.63} & 3.57 & 2.55 & \underline{2.96} & \textbf{2.91} \\
BizSage (Iteration 2) & \underline{3.92} & \underline{2.88} & \underline{3.38} & \underline{2.99} & \underline{4.21} & \underline{2.97} & \underline{2.82} & \textbf{3.78} & \underline{3.81} & \underline{2.76} & 2.94 & \underline{2.90} \\
\rowcolor{oursrow} BizSage (Iteration 3) & \textbf{3.94} & \textbf{2.96} & \textbf{3.41} & \textbf{3.01} & \textbf{4.25} & \textbf{2.99} & \textbf{2.86} & \textbf{3.78} & \textbf{4.02} & \textbf{2.95} & \textbf{3.07} & 2.86 \\
\midrule
\multicolumn{13}{@{}l}{\textbf{\textit{DeepSeek-V4-Flash}}} \\
BizSage (Iteration 1) & 4.13 & \underline{2.92} & 3.47 & 3.42 & \underline{4.22} & \underline{1.98} & 2.64 & 2.77 & 4.18 & 3.06 & 3.33 & 3.21 \\
BizSage (Iteration 2) & \underline{4.15} & 2.91 & \underline{3.52} & \underline{3.43} & 4.21 & 1.93 & \underline{2.85} & \underline{2.94} & \underline{4.29} & \underline{3.11} & \underline{3.37} & \underline{3.26} \\
\rowcolor{oursrow} BizSage (Iteration 3) & \textbf{4.65} & \textbf{2.96} & \textbf{3.89} & \textbf{4.08} & \textbf{4.74} & \textbf{3.07} & \textbf{3.06} & \textbf{3.79} & \textbf{4.46} & \textbf{3.88} & \textbf{3.93} & \textbf{3.70} \\
\doublerule
\end{tabular}
\caption{Self-evolution across three strategy-refinement iterations per backbone. Each iteration refines strategies using 10 development queries. \textbf{Bold} and \underline{underline} mark the best and second-best scores per column within each backbone.}
\label{tab:self-evolution}
\end{table*}

Table~\ref{tab:self-evolution} tracks quality scores across three strategy-refinement iterations for all three backbones.
Iteration 3 achieves the best score on nearly every metric and backbone.
Since Iteration 1 runs with an empty strategy memory $M_Q$, the gap between Iteration 1 and Iteration 3 also serves as a direct ablation of the cross-task strategy memory.
Grounding shows the largest gains: it starts as the lowest-scoring dimension in Iteration 1, and the gap between Grounding and other dimensions narrows substantially after three iterations, indicating that self-evolution preferentially repairs the dimension with the largest initial deficit.
The three backbones exhibit distinct improvement trajectories: GPT-5.5 improves evenly across iterations with the highest final scores yet not the largest total gain, suggesting that a high starting point compresses the room for improvement; DeepSeek-V4-Flash stalls for the first two iterations before producing the single largest per-iteration jump across all backbones with the greatest total gain, suggesting a threshold effect in which accumulated strategies only take effect once a critical mass is reached; Gemini-3-Flash shows the smallest total gain with several metrics declining slightly, indicating that strategies are difficult to execute when backbone capacity is insufficient.
Overall, self-evolution reshapes the quality profile by \textbf{concentrating its corrections on the weakest dimension rather than lifting all dimensions uniformly}, and its lower bound of effectiveness is determined by whether the backbone can reliably execute the accumulated strategies.

\begin{table*}[t]
\centering
\footnotesize
\setlength{\tabcolsep}{3.6pt}
\renewcommand{\arraystretch}{1.22}
\begin{tabular}{@{}l|cccc|cccc|cccc@{}}
\doublerule
 & \multicolumn{4}{c|}{Survey} & \multicolumn{4}{c|}{Idea Formulation} & \multicolumn{4}{c}{Research Plan} \\
\cmidrule(lr){2-5} \cmidrule(lr){6-9} \cmidrule(lr){10-13}
\multirow{-2}{*}{\textbf{Method}} & Rel. & Grd. & Cov. & Syn. & Rel. & Grd. & Nov. & Fea. & Rel. & Grd. & Rig. & Fea. \\
\midrule
\rowcolor{oursrow} BizSage (full) & \textbf{4.77} & \textbf{4.14} & \textbf{4.04} & \textbf{3.99} & \textbf{4.71} & \textbf{3.42} & \textbf{3.43} & \textbf{3.95} & \textbf{4.89} & \textbf{4.36} & \textbf{4.07} & \textbf{3.95} \\
\midrule
\multicolumn{13}{@{}l}{\textit{Ablations}} \\
\hspace{1em}$\vdash$ without Section-Level Knowledge Graph & \underline{4.32} & 3.19 & \underline{3.99} & 3.92 & \underline{4.48} & 2.26 & \underline{3.35} & \underline{3.61} & \underline{4.73} & \underline{3.56} & \underline{4.01} & \underline{3.55} \\
\hspace{1em}$\vdash$ without Reviewer & 4.16 & \underline{4.06} & 3.89 & 3.83 & 4.36 & \underline{3.08} & \underline{3.35} & 3.50 & 4.62 & 3.27 & 3.97 & 3.32 \\
\hspace{1em}$\vdash$ without Discussant & 4.20 & 4.03 & 3.79 & 3.68 & 4.26 & 3.07 & 3.19 & 3.36 & 4.57 & 3.26 & 3.87 & 3.18 \\
\hspace{1em}$\vdash$ without Iterative Refinement & 4.23 & 3.15 & 3.93 & \underline{3.94} & 4.36 & 2.23 & 3.34 & 3.55 & 4.71 & 2.41 & 3.97 & 3.46 \\
\doublerule
\end{tabular}
\caption{Ablation study on the GPT-5.5 backbone. Each variant removes one component from the full system. \textbf{Bold} and \underline{underline} mark the best and second-best scores per column.}
\label{tab:ablation}
\end{table*}

\subsection{Ablation Study (RQ3)}

To isolate each component's contribution, we ablate four modules from the full system using GPT-5.5, with results reported in Table~\ref{tab:ablation}.
The four components divide along two axes, citation accuracy and content quality: Section-Level Knowledge Graph and Iterative Refinement primarily affect Grounding, while Reviewer and Discussant primarily affect Coverage and Synthesis.
For citation accuracy, removing Iterative Refinement causes a larger Grounding drop than removing Section-Level Knowledge Graph, indicating that complex citation structures require multi-round correction rather than merely better retrieval input.
Notably, removing the knowledge graph leaves Coverage largely intact while Grounding drops sharply, suggesting that the bottleneck lies in attribution precision rather than literature breadth.
For content quality, the Discussant has a greater impact than the Reviewer, because the former proactively injects alternative perspectives during generation while the latter only scores existing output.
Overall, the four components exhibit a \textbf{largely complementary impact pattern}: retrieval modules and collaborative agents address largely independent failure modes, and strengthening one cannot compensate for deficiencies in the other.

\section{Discussion}
\subsection{Retrieval Efficiency Analysis}

We compare BizSage with four retrieval-only variants and three agentic
framework baselines.
LKG uses the corpus-level graph without additional graph diffusion,
while Long-Walk PPR follows Personalized PageRank-style propagation over the
graph~\citep{page1999pagerank, haveliwala2002topic, 10.1109/TKDE.2024.3376000}.
Katz Diffusion~\citep{Katz1953ANS} and Heat-Kernel
Diffusion~\citep{Kondor2002DiffusionKO} replace PPR with alternative graph
diffusion operators.
The agentic baselines include ResearchAgent~\citep{baek2025researchagent},
Idea2Story~\citep{xu2026idea2story}, and AI Co-Scientist~\citep{gottweis2025towards}.

\begin{figure*}[t]
\centering
\includegraphics[width=\textwidth]{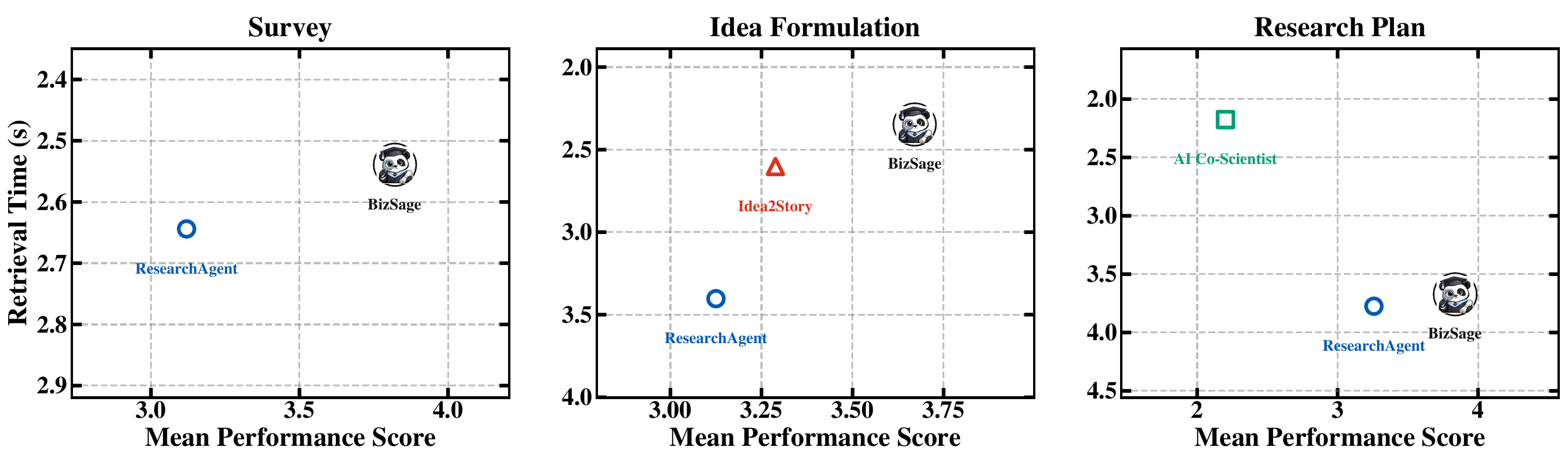}
\caption{Efficiency--performance matrix across the three task types. Each point represents one method; the x-axis shows mean quality score and the y-axis shows retrieval time in seconds. Methods closer to the lower-right corner achieve higher quality with shorter retrieval time.}
\label{fig:efficiency-matrix}
\end{figure*}

Figure~\ref{fig:efficiency-matrix} summarizes the efficiency and performance trade-off.
BizSage occupies the favorable high-quality, low-time region for survey and idea
formulation, combining the highest mean quality with the shortest retrieval
time among the plotted methods.
For research plan, AI Co-Scientist retrieves faster but has substantially
lower quality, while BizSage remains close to ResearchAgent in retrieval time
and yields a much stronger quality score.
The full runtime table in Appendix~\ref{app:efficiency-runtime} further shows
that diffusion-heavy retrieval variants, especially Long-Walk PPR and
Heat-Kernel Diffusion, incur larger propagation costs than BizSage's default
configuration.

\subsection{Knowledge Graph Construction Analysis}

\begin{table}[t]
\centering
\footnotesize
\setlength{\tabcolsep}{6pt}
\begin{tabular}{@{}l|ccc@{}}
\doublerule
\noalign{\vspace{2pt}}
\textbf{Extraction Model} & \textbf{NV} & \textbf{RS} & \textbf{MC} \\
\midrule
Qwen3-8B & 65.0 & 44.4 & 42.9 \\
GPT-4o & 90.0 & 88.2 & 92.3 \\
GPT-5.5 & 95.2 & 94.4 & 93.3 \\
\doublerule
\end{tabular}
\caption{Quality of section-level graph extraction across extraction models on 50 manually annotated paper sections. NV = Node Validity, RS = Relation Support, MC = Merge Consistency. All values are percentages, and higher is better.}
\label{tab:extraction-models}
\end{table}

Building section-level graphs for the full corpus with GPT-4o cost approximately \$7{,}300 in API usage.
This expense is incurred once at ingestion and is absorbed by the operator of the system; it is not repeated per query and is not passed on to users.
The released graph covers publications dating back to the 1920s, and bringing it up to date only requires extracting and merging the papers added since the previous build, a workload that is small relative to the initial construction.

Selecting GPT-4o for extraction balances graph quality against expense.
To quantify this trade-off, we collect 50 paper sections with manually annotated entities and relations and run Qwen3-8B, GPT-4o, and GPT-5.5 under an identical prompt and merging pipeline.
Three measures are reported in Table~\ref{tab:extraction-models}: node validity, the proportion of extracted nodes that denote a genuine concept in the annotated section; relation support, the proportion of extracted edges that the passage directly asserts or clearly entails; and merge consistency, the proportion of cross-document merges that the annotators confirm as denoting one and the same entity.
GPT-5.5 exceeds GPT-4o by 5.2 points on node validity and 6.2 points on relation support, with the gain arising chiefly from more complete relation extraction in a few long sections, while merge consistency differs by only one point.
Given the far higher inference cost and latency of GPT-5.5, applying it to the full corpus would not be cost-effective.
Qwen3-8B produces unstable extractions on this task, and its errors propagate and compound during node merging, so it is unsuitable for building a corpus-scale graph.

\subsection{Citation Integrity Analysis}

\begin{table}[t]
\centering
\footnotesize
\setlength{\tabcolsep}{6pt}
\begin{tabular}{@{}l|cc@{}}
\doublerule
\noalign{\vspace{2pt}}
\textbf{Method} & \textbf{\#Citations} & \textbf{Hallucination Rate} \\
\midrule
\textbf{BizSage (Ours)} & 894 & 0.0\% \\
InternAgent-1.5 & 98 & 5.1\% \\
EvoScientist & 75 & 2.7\% \\
\cmidrule{1-3}
AI-Scientist-v2 & 0 & -- \\
ResearchAgent & 0 & -- \\
\cmidrule{1-3}
Idea2Story$^{*}$ & 0 & -- \\
AI Co-Scientist$^{*}$ & 0 & -- \\
\doublerule
\end{tabular}
\caption{Citation integrity on the survey task. $^{*}$These systems do not produce surveys; we evaluate twenty outputs from their native tasks instead.}
\label{tab:citation-integrity}
\end{table}

Reliable citations allow researchers to verify claims in system-generated text against published literature.
Their absence leaves researchers unable to assess which claims are substantiated, while fabricated citations mislead by lending false credibility.
We select the survey task for its intensive citation demands, extract all citations from twenty outputs per system, and verify each against CrossRef\footnote{\url{https://www.crossref.org/}} with human review of unmatched entries.
This audit covers whether each cited work exists, whether its bibliographic details are accurate, and whether it is topically relevant to the queried domain.

As shown in Table~\ref{tab:citation-integrity}, BizSage produces zero hallucinated citations across 894 entries, while InternAgent-1.5 and EvoScientist contain five and two hallucinated entries respectively, with approximately 40\% of InternAgent-1.5's valid citations drawn from unrelated fields.
The remaining systems produce no citations at all, meaning their outputs lack verifiable evidence trail for academic writing.
Beyond correctness, Appendix~\ref{app:citation-concentration} shows that BizSage's citations are widely dispersed across the corpus rather than concentrated on a small set of papers.

Prior work has shown that standalone LLMs hallucinate citations at rates of 14\% to 95\%~\cite{linardon2025citationfabrication,xu2026ghostcite,sakai2026hallucitation}.
This analysis extends that concern to research agent systems and identifies domain drift, where citations refer to real but topically irrelevant publications, as a failure mode that current hallucination detection cannot capture.
Addressing both hallucination and drift requires domain-aligned retrieval with explicit verification at the output stage.

\section{Conclusion}

We present BizSage, a self-evolving multi-agent framework for automated research in economics and business, which leverages a corpus-level LKG with PPR retrieval to support three progressive tasks: survey, idea formulation, and research plan.
On a benchmark spanning four sub-domains and three backbone LLMs, BizSage achieves leading scores on the majority of evaluation dimensions, with zero hallucinated citations in a sampled verification study.
Our analysis further reveals two findings relevant to future system design: self-evolution concentrates its corrections on the weakest quality dimension rather than lifting all dimensions uniformly, and retrieval modules and collaborative agents address largely independent failure modes, making both indispensable for domain-specific research quality.
We hope this work provides a foundation for extending automated research assistance to broader social-science disciplines.

\section{Limitations}

Despite the promising results demonstrated in this study, several limitations must be acknowledged.
The source corpus used for knowledge graph construction is drawn from 73 peer-reviewed journals and does not include working papers, conference proceedings, or preprints, potentially missing the most recent findings in fast-moving research areas.
In addition, the system and all evaluation materials are in English, and the applicability to non-English academic literature remains unexplored.
The LKG is built once during corpus ingestion and cannot automatically incorporate newly published work, requiring periodic reconstruction to maintain up-to-date coverage.
The self-evolution mechanism has only been tested for three iterations, and the long-term convergence behavior as well as whether accumulated strategies may become mutually conflicting have not been studied. Finally, the benchmark contains only 300 test queries. Since each query is retained through strict expert curation and elicits a long, citation-dense output that is costly to generate and judge, this scale already supports a reliable comparison across systems, and we will consider larger-scale evaluation in future work.

\section*{Acknowledgments}
We thank the anonymous reviewers for their valuable comments and suggestions.

\bibliography{acl_2025}
\clearpage
\appendix

\definecolor{rowshade}{HTML}{EDF2F7}
\newcommand{\jurl}[2]{\href{https://#1}{#2}}

\section{Source Journal Corpus}
\label{app:journals}

We construct a domain corpus from peer-reviewed journals in four fields
central to economics and business research: economics, finance, operations
management, and statistics.
Journal selection follows a two-step process.
First, we anchor each domain on its widely acknowledged flagship
journals, identified through bibliometric ranking
studies~\citep{currie2020finance, kalaitzidakis2011updated, petersen2011journal}.
Second, we extend coverage by adding leading field journals that hold
top-tier ratings in the Chartered Association of Business Schools (ABS)
Academic Journal Guide (AJG)~\citep{walker2021methodology, hudson2025does}.
The resulting journal list comprises 73 journals.
The processed corpus snapshot used for knowledge graph (KG) construction contains
134{,}234 papers spanning publication dates from the 1920s to 2025.
It includes 56{,}213 economics papers, 40{,}466 finance papers, 8{,}094
operations management papers, and 29{,}461 statistics papers.
Tables~\ref{tab:econ-journals}--\ref{tab:stat-journals} list all
journals with their publishers, temporal coverage, processed paper
counts, and access uniform resource locators (URLs).
Publisher abbreviations in the journal tables use official short names:
American Economic Association (AEA), Oxford University Press (OUP),
Massachusetts Institute of Technology (MIT) Press, University of Wisconsin
(U Wisconsin), Econometric Society (Econometric Soc.), and the Institute for
Operations Research and the Management Sciences (INFORMS).

\subsection{Economics}
\label{app:econ}

Economics research values methodological rigour, disciplinary reach, and
balanced field coverage.
Bibliometric surveys consistently place a small set of flagship general
journals at the centre of citation influence and promotion
decisions~\citep{kalaitzidakis2011updated, heckman2020publishing};
our corpus includes four of these:
\emph{American Economic Review}, \emph{Quarterly Journal of Economics},
\emph{Econometrica}, and \emph{Review of Economic Studies}.
To avoid overweighting these flagships, we add the top-ranked field
journals that remain in the overall top-30 after field-normalized and
multi-criteria
rankings~\citep{combes2010inferring, engemann2009journal, ritzberger2008ranking},
covering econometrics (\emph{Journal of Econometrics},
\emph{Econometric Theory}),
development (\emph{Journal of Development Economics}),
and labor (\emph{Journal of Human Resources}).
The four \emph{American Economic Journal} sub-titles, launched in 2009,
provide additional coverage of applied, macro, micro, and policy
research.
\emph{Experimental Economics} ensures representation of laboratory and
behavioural methods, and \emph{Review of Economics and Statistics}
bridges empirical economics and quantitative social science.
The selected 22 journals and their details are shown in
Table~\ref{tab:econ-journals}.

\begin{table*}[t]
  \centering
  \setlength{\tabcolsep}{3.5pt}
  \footnotesize
  \begin{NiceTabular}{@{}
    >{\raggedright\arraybackslash}p{5.80cm}
    >{\centering\arraybackslash}p{2.45cm}
    >{\centering\arraybackslash}p{1.50cm}
    >{\raggedright\arraybackslash}p{3.55cm}
    >{\raggedleft\arraybackslash}p{1.25cm}
  @{}}[cell-space-limits=3pt]
  \CodeBefore
    \rowcolors{2}{rowshade}{white}
  \Body
  \toprule
  \rowcolor{white}
  \textbf{Journal} & \textbf{Publisher} & \textbf{Coverage} & \textbf{Website} & \textbf{Papers} \\
  \midrule
  American Economic Review                         & AEA              & 1980--2021 & \jurl{aeaweb.org/journals/aer}{aeaweb.org/j/aer} & 8{,}760 \\
  Journal of Econometrics                          & Elsevier         & 1973--2022 & \jurl{www.sciencedirect.com/journal/journal-of-econometrics}{elsevier.com/j/econometrics} & 5{,}637 \\
  Journal of Economic Theory                       & Elsevier         & 1974--2022 & \jurl{www.sciencedirect.com/journal/journal-of-economic-theory}{elsevier.com/j/econ-theory} & 5{,}151 \\
  The Economic Journal                             & OUP              & 1927--2023 & \jurl{academic.oup.com/ej}{oup.com/ej} & 4{,}832 \\
  Econometrica                                     & Econometric Soc. & 1980--2023 & \jurl{www.econometricsociety.org/publications/econometrica}{econometricsociety.org} & 4{,}622 \\
  Review of Economics and Statistics               & MIT Press        & 1969--2018 & \jurl{direct.mit.edu/rest}{direct.mit.edu/rest} & 3{,}744 \\
  Review of Economic Studies                       & OUP              & 1936--2016 & \jurl{academic.oup.com/restud}{oup.com/restud} & 3{,}309 \\
  Journal of Development Economics                 & Elsevier         & 1975--2015 & \jurl{www.sciencedirect.com/journal/journal-of-development-economics}{elsevier.com/j/dev-econ} & 3{,}207 \\
  International Economic Review                    & Wiley            & 1961--2020 & \jurl{onlinelibrary.wiley.com/journal/14682354}{wiley.com/j/14682354} & 3{,}116 \\
  Quarterly Journal of Economics                   & OUP              & 1927--2021 & \jurl{academic.oup.com/qje}{oup.com/qje} & 2{,}421 \\
  Journal of Human Resources                       & U Wisconsin      & 1966--2018 & \jurl{jhr.uwpress.org}{jhr.uwpress.org} & 1{,}892 \\
  Econometric Theory                               & Cambridge        & 1985--2023 & \jurl{www.cambridge.org/core/journals/econometric-theory}{cambridge.org/j/ect} & 1{,}785 \\
  RAND Journal of Economics                        & Wiley            & 1973--2017 & \jurl{onlinelibrary.wiley.com/journal/17562171}{wiley.com/j/17562171} & 1{,}593 \\
  Journal of the European Economic Association     & OUP              & 2003--2018 & \jurl{academic.oup.com/jeea}{oup.com/jeea} & 1{,}310 \\
  Experimental Economics                           & Springer         & 2000--2023 & \jurl{link.springer.com/journal/10683}{springer.com/j/10683} & 828 \\
  American Economic Journal: Economic Policy       & AEA              & 2009--2021 & \jurl{aeaweb.org/journals/pol}{aeaweb.org/j/pol} & 821 \\
  American Economic Journal: Microeconomics        & AEA              & 2009--2021 & \jurl{aeaweb.org/journals/mic}{aeaweb.org/j/mic} & 670 \\
  American Economic Journal: Macroeconomics        & AEA              & 2009--2021 & \jurl{aeaweb.org/journals/mac}{aeaweb.org/j/mac} & 668 \\
  Theoretical Economics                            & Econometric Soc. & 2006--2020 & \jurl{econtheory.org}{econtheory.org} & 556 \\
  American Economic Journal: Applied Economics     & AEA              & 2009--2021 & \jurl{aeaweb.org/journals/app}{aeaweb.org/j/app} & 507 \\
  Quantitative Economics                           & Econometric Soc. & 2010--2022 & \jurl{qeconomics.org}{qeconomics.org} & 407 \\
  Journal of Economic Growth                       & Springer         & 1996--2023 & \jurl{link.springer.com/journal/10887}{springer.com/j/10887} & 377 \\
  \bottomrule
  \end{NiceTabular}
  \caption{Economics journals in the source corpus.}
  \label{tab:econ-journals}
\end{table*}

\subsection{Finance}
\label{app:finance}

Finance scholarship values methodological rigour, cross-field impact,
and sub-field balance.
Citation and promotion studies identify the ``Top Three'' general
journals---\emph{Journal of Finance}, \emph{Journal of Financial Economics},
and \emph{Review of Financial Studies}---as dominant in career-making
citations~\citep{currie2011finance, currie2020finance, borokhovich2000analysis}.
We extend coverage with leading field journals that rank A/A$^{+}$ in
active-scholar assessments~\citep{borokhovich2011framework}, spanning
corporate finance (\emph{Journal of Corporate Finance}),
market microstructure (\emph{Journal of Financial Markets},
\emph{Journal of Financial Intermediation}),
and risk management (\emph{Journal of Financial Stability}).
Insurance and actuarial science are represented by
\emph{Insurance: Mathematics and Economics}, \emph{ASTIN Bulletin},
\emph{Journal of Risk and Insurance}, and two Scandinavian and North
American actuarial outlets.
Real-estate finance is covered by \emph{Real Estate Economics} and
\emph{Journal of Real Estate Finance and Economics}.
The resulting 29-journal set captures the citation core of contemporary
financial economics, as shown in Table~\ref{tab:fin-journals}.

\begin{table*}[!p]
  \centering
  \setlength{\tabcolsep}{3.5pt}
  \footnotesize
  \begin{NiceTabular}{@{}
    >{\raggedright\arraybackslash}p{5.80cm}
    >{\centering\arraybackslash}p{2.45cm}
    >{\centering\arraybackslash}p{1.50cm}
    >{\raggedright\arraybackslash}p{3.55cm}
    >{\raggedleft\arraybackslash}p{1.25cm}
  @{}}[cell-space-limits=3pt]
  \CodeBefore
    \rowcolors{2}{rowshade}{white}
  \Body
  \toprule
  \rowcolor{white}
  \textbf{Journal} & \textbf{Publisher} & \textbf{Coverage} & \textbf{Website} & \textbf{Papers} \\
  \midrule
  Journal of Finance                               & Wiley            & 1949--2020 & \jurl{onlinelibrary.wiley.com/journal/15406261}{wiley.com/j/15406261} & 5{,}902 \\
  Journal of Financial Economics                   & Elsevier         & 1978--2022 & \jurl{www.sciencedirect.com/journal/journal-of-financial-economics}{elsevier.com/j/jfe} & 3{,}201 \\
  Review of Financial Studies                      & OUP              & 1988--2021 & \jurl{academic.oup.com/rfs}{oup.com/rfs} & 2{,}529 \\
  Journal of International Money and Finance       & Elsevier         & 1982--2023 & \jurl{www.sciencedirect.com/journal/journal-of-international-money-and-finance}{elsevier.com/j/jimf} & 2{,}264 \\
  Journal of Corporate Finance                     & Elsevier         & 1997--2021 & \jurl{www.sciencedirect.com/journal/journal-of-corporate-finance}{elsevier.com/j/jcf} & 2{,}214 \\
  Insurance: Mathematics and Economics             & Elsevier         & 1982--2022 & \jurl{www.sciencedirect.com/journal/insurance-mathematics-and-economics}{elsevier.com/j/ime} & 1{,}806 \\
  Financial Review                                 & Wiley            & 1968--2019 & \jurl{onlinelibrary.wiley.com/journal/15406288}{wiley.com/j/15406288} & 1{,}598 \\
  Journal of Applied Corporate Finance             & Wiley            & 1988--2024 & \jurl{onlinelibrary.wiley.com/journal/17456622}{wiley.com/j/17456622} & 1{,}595 \\
  Review of Quantitative Finance and Accounting    & Springer         & 1992--2024 & \jurl{link.springer.com/journal/11156}{springer.com/j/11156} & 1{,}550 \\
  ASTIN Bulletin                                   & Cambridge        & 1960--2024 & \jurl{www.cambridge.org/core/journals/astin-bulletin}{cambridge.org/j/astin} & 1{,}521 \\
  Corporate Governance: An International Review    & Wiley            & 1993--2022 & \jurl{onlinelibrary.wiley.com/journal/14678683}{wiley.com/j/14678683} & 1{,}485 \\
  Journal of Financial Research                    & Wiley            & 1978--2015 & \jurl{onlinelibrary.wiley.com/journal/14756803}{wiley.com/j/14756803} & 1{,}424 \\
  Real Estate Economics                            & Wiley            & 1973--2023 & \jurl{onlinelibrary.wiley.com/journal/15406229}{wiley.com/j/15406229} & 1{,}421 \\
  European Journal of Finance                      & Taylor \& Francis & 1997--2025 & \jurl{www.tandfonline.com/toc/rejf20/current}{tandfonline.com/rejf20} & 1{,}398 \\
  Applied Financial Economics                      & Taylor \& Francis & 1997--2012 & \jurl{www.tandfonline.com/toc/rafe20/current}{tandfonline.com/rafe20} & 1{,}136 \\
  North American Actuarial Journal                 & Taylor \& Francis & 1997--2024 & \jurl{www.tandfonline.com/toc/uaaj20/current}{tandfonline.com/uaaj20} & 1{,}098 \\
  European Financial Management                    & Wiley            & 1995--2016 & \jurl{onlinelibrary.wiley.com/journal/1468036x}{wiley.com/j/1468036x} & 1{,}089 \\
  Journal of Financial Stability                   & Elsevier         & 2005--2020 & \jurl{www.sciencedirect.com/journal/journal-of-financial-stability}{elsevier.com/j/jfs} & 1{,}075 \\
  Journal of Risk and Insurance                    & Wiley            & 1956--2020 & \jurl{onlinelibrary.wiley.com/journal/15396975}{wiley.com/j/15396975} & 949 \\
  Mathematical Finance                             & Wiley            & 1991--2018 & \jurl{onlinelibrary.wiley.com/journal/14679965}{wiley.com/j/14679965} & 884 \\
  Review of Finance                                & OUP              & 1997--2023 & \jurl{academic.oup.com/rof}{oup.com/rof} & 843 \\
  Journal of Financial Services Research           & Springer         & 1988--2023 & \jurl{link.springer.com/journal/10693}{springer.com/j/10693} & 644 \\
  International Review of Finance                  & Wiley            & 2000--2017 & \jurl{onlinelibrary.wiley.com/journal/14682443}{wiley.com/j/14682443} & 616 \\
  Journal of Financial Intermediation              & Elsevier         & 2001--2017 & \jurl{www.sciencedirect.com/journal/journal-of-financial-intermediation}{elsevier.com/j/jfi} & 610 \\
  Journal of Financial Markets                     & Elsevier         & 2003--2020 & \jurl{www.sciencedirect.com/journal/journal-of-financial-markets}{elsevier.com/j/finmar} & 584 \\
  Scandinavian Actuarial Journal                   & Taylor \& Francis & 1997--2024 & \jurl{www.tandfonline.com/toc/sact20/current}{tandfonline.com/sact20} & 528 \\
  Review of Asset Pricing Studies                  & OUP              & 2011--2020 & \jurl{academic.oup.com/raps}{oup.com/raps} & 204 \\
  Review of Corporate Finance Studies              & OUP              & 2012--2021 & \jurl{academic.oup.com/rcfs}{oup.com/rcfs} & 191 \\
  Journal of Real Estate Finance and Economics     & Springer         & 2007--2023 & \jurl{link.springer.com/journal/11146}{springer.com/j/11146} & 107 \\
  \bottomrule
  \end{NiceTabular}
  \caption{Finance journals in the source corpus.}
  \label{tab:fin-journals}
\end{table*}

\subsection{Operations Management}
\label{app:om}

Operations management and operations research are organized around a
compact citation core.
Bibliometric analyses show that a small number of INFORMS-published journals
account for a large share of within-field
citations~\citep{petersen2011journal, song2022scientific}, and these
journals hold top-tier AJG 2021 ratings~\citep{walker2021methodology}.
We anchor the corpus on four INFORMS flagships---\emph{Management
Science}, \emph{Operations Research}, \emph{Transportation Science},
and \emph{INFORMS Journal on Applied Analytics}---which collectively
represent the core of the discipline.
To ensure breadth in methodology and applications, we add journals
covering production systems (\emph{International Journal of Production
Research}, \emph{Journal of Operations Management}),
mathematical optimization (\emph{Mathematical Programming},
\emph{Journal of Optimization Theory and Applications}),
scheduling and logistics (\emph{Journal of Scheduling},
\emph{Naval Research Logistics}, \emph{Operations Research Letters}),
and decision analysis (\emph{Decision Sciences}).
\emph{Institute of Industrial and Systems Engineers (IISE) Transactions}
(formerly \emph{Institute of Industrial Engineers (IIE) Transactions}) provides
additional coverage of industrial and systems engineering methods.
The selected 13 journals are listed in Table~\ref{tab:om-journals}.

\begin{table*}[!p]
  \centering
  \setlength{\tabcolsep}{3.5pt}
  \footnotesize
  \begin{NiceTabular}{@{}
    >{\raggedright\arraybackslash}p{5.80cm}
    >{\centering\arraybackslash}p{2.45cm}
    >{\centering\arraybackslash}p{1.50cm}
    >{\raggedright\arraybackslash}p{3.55cm}
    >{\raggedleft\arraybackslash}p{1.25cm}
  @{}}[cell-space-limits=3pt]
  \CodeBefore
    \rowcolors{2}{rowshade}{white}
  \Body
  \toprule
  \rowcolor{white}
  \textbf{Journal} & \textbf{Publisher} & \textbf{Coverage} & \textbf{Website} & \textbf{Papers} \\
  \midrule
  INFORMS Journal on Applied Analytics             & INFORMS          & 1971--2023 & \jurl{pubsonline.informs.org/journal/inte}{informs.org/j/inte} & 3{,}204 \\
  Operations Research                              & INFORMS          & 1976--2024 & \jurl{pubsonline.informs.org/journal/opre}{informs.org/j/opre} & 2{,}164 \\
  Transportation Science                           & INFORMS          & 1967--2011 & \jurl{pubsonline.informs.org/journal/trsc}{informs.org/j/trsc} & 1{,}328 \\
  Management Science                               & INFORMS          & 1970--1977 & \jurl{pubsonline.informs.org/journal/mnsc}{informs.org/j/mnsc} & 1{,}043 \\
  Mathematical Programming                         & Springer         & 1995--2024 & \jurl{link.springer.com/journal/10107}{springer.com/j/10107} & 70 \\
  Operations Research Letters                      & Elsevier         & 2004--2023 & \jurl{www.sciencedirect.com/journal/operations-research-letters}{elsevier.com/j/orl} & 38 \\
  International Journal of Production Research     & Taylor \& Francis & 1980--2023 & \jurl{www.tandfonline.com/toc/tprs20/current}{tandfonline.com/tprs20} & 37 \\
  Decision Sciences                                & Wiley            & 1998--2011 & \jurl{onlinelibrary.wiley.com/journal/15405915}{wiley.com/j/15405915} & 37 \\
  Journal of Optimization Theory and Applications  & Springer         & 2007--2022 & \jurl{link.springer.com/journal/10957}{springer.com/j/10957} & 36 \\
  Naval Research Logistics                         & Wiley            & 2001--2018 & \jurl{onlinelibrary.wiley.com/journal/15206750}{wiley.com/j/15206750} & 35 \\
  Journal of Scheduling                            & Springer         & 2007--2022 & \jurl{link.springer.com/journal/10951}{springer.com/j/10951} & 35 \\
  Journal of Operations Management                 & Wiley            & 1983--2016 & \jurl{onlinelibrary.wiley.com/journal/18731317}{wiley.com/j/18731317} & 35 \\
  IISE Transactions                                & Taylor \& Francis & 1973--2023 & \jurl{www.tandfonline.com/toc/uiie20/current}{tandfonline.com/uiie20} & 32 \\
  \bottomrule
  \end{NiceTabular}
  \caption{Operations management journals in the source corpus.}
  \label{tab:om-journals}
\end{table*}

\subsection{Statistics}
\label{app:stat}

Citation capital in statistics is concentrated: the ten most-cited
statistics journals capture just over half of all within-field
citations~\citep{wang2022statistics}, and field-normalized analyses
reveal pronounced inequality in citation
distributions~\citep{crespo2013measurement, ruiz2015field,  kozlowski2024decrease}.
We select journals that cover the field's core areas: mathematical
statistics and probability theory (\emph{Biometrika},
\emph{Journal of the Royal Statistical Society (JRSS) Series~B}), computational and graphical methods
(\emph{Journal of Computational and Graphical Statistics}),
applied and public-health statistics (\emph{Biostatistics},
\emph{JRSS Series~C}), and a generalist forum
(\emph{Scandinavian Journal of Statistics}).
\emph{Journal of Business \& Economic Statistics} is included in this
domain rather than economics, as it bridges econometrics and
statistical methodology and is frequently cited by both communities.
\emph{Computational Statistics \& Data Analysis} and \emph{Journal of
Multivariate Analysis} round out the computational and theoretical
coverage respectively.
The selected 9 journals are listed in Table~\ref{tab:stat-journals}.

\begin{table*}[!p]
  \centering
  \setlength{\tabcolsep}{3.5pt}
  \footnotesize
  \begin{NiceTabular}{@{}
    >{\raggedright\arraybackslash}p{5.80cm}
    >{\centering\arraybackslash}p{2.45cm}
    >{\centering\arraybackslash}p{1.50cm}
    >{\raggedright\arraybackslash}p{3.55cm}
    >{\raggedleft\arraybackslash}p{1.25cm}
  @{}}[cell-space-limits=3pt]
  \CodeBefore
    \rowcolors{2}{rowshade}{white}
  \Body
  \toprule
  \rowcolor{white}
  \textbf{Journal} & \textbf{Publisher} & \textbf{Coverage} & \textbf{Website} & \textbf{Papers} \\
  \midrule
  Biometrika                                       & OUP              & 1925--2019 & \jurl{academic.oup.com/biomet}{oup.com/biomet} & 6{,}947 \\
  Computational Statistics \& Data Analysis        & Elsevier         & 1990--2022 & \jurl{www.sciencedirect.com/journal/computational-statistics-and-data-analysis}{elsevier.com/j/csda} & 4{,}906 \\
  Journal of Multivariate Analysis                 & Elsevier         & 1976--2022 & \jurl{www.sciencedirect.com/journal/journal-of-multivariate-analysis}{elsevier.com/j/jmva} & 4{,}791 \\
  Journal of Business \& Economic Statistics       & Taylor \& Francis & 1983--2024 & \jurl{www.tandfonline.com/toc/ubes20/current}{tandfonline.com/ubes20} & 3{,}418 \\
  JRSS Series~C (Applied Statistics)               & OUP              & 1950--2019 & \jurl{academic.oup.com/jrsssc}{oup.com/jrsssc} & 3{,}411 \\
  JRSS Series~B (Statistical Methodology)          & OUP              & 1936--2010 & \jurl{academic.oup.com/jrsssb}{oup.com/jrsssb} & 1{,}717 \\
  Journal of Computational and Graphical Statistics & Taylor \& Francis & 1992--2025 & \jurl{www.tandfonline.com/toc/ucgs20/current}{tandfonline.com/ucgs20} & 1{,}659 \\
  Scandinavian Journal of Statistics               & Wiley            & 1974--2017 & \jurl{onlinelibrary.wiley.com/journal/14679469}{wiley.com/j/14679469} & 1{,}442 \\
  Biostatistics                                    & OUP              & 2000--2022 & \jurl{academic.oup.com/biostatistics}{oup.com/biostatistics} & 1{,}170 \\
  \bottomrule
  \end{NiceTabular}
  \caption{Statistics journals in the source corpus.}
  \label{tab:stat-journals}
\end{table*}

\section{Related Works}
\label{sec:related}

\subsection{Multi-Agent Systems for Scientific Research}

Multi-agent systems for scientific research have advanced along two
directions: idea generation through propose-critique-refine
cycles~\citep{baek2025researchagent, su2025many, gottweis2025towards}, and
end-to-end automation of the research
workflow~\citep{lu2024ai, yamada2025ai, schmidgall2025agent,
schmidgall2025agentrxiv, tang2025ai, team2025internagent}.
Recent work also addresses cross-task capability
retention~\citep{weidener2026rethinking, bicker2026aster, lin2026scider}:
EvoScientist~\citep{lyu2026evoscientist} maintains persistent ideation and
experimentation memories for improvement across tasks,
DeepScientist~\citep{weng2025deepscientist} retains findings for
long-horizon discovery, and
InternAgent-1.5~\citep{feng2026internagent} provides a unified framework for
extended research trajectories.
Despite these advances, existing systems and their evaluation
benchmarks~\citep{luo2025benchmarking, liu2025hypobench} target
computational disciplines where experiments are validated through code
execution.
In economics and business research, by contrast, current efforts are limited
to agentic workflow tools~\citep{dawid2025agentic} and methodological
guidelines~\citep{korinek2025ai}; social science applications have explored
only synthetic-agent settings~\citep{ji2026leveraging}.
This gap reflects challenges in retrieval granularity and quality adaptation
that existing architectures do not address.

\subsection{Knowledge Retrieval for Scientific Research}

Retrieval-augmented generation (RAG)~\citep{lewis2020retrieval} has become the
standard paradigm for grounding LLM outputs in external knowledge, with
dense retrieval methods~\citep{karpukhin2020dense, izacard2021unsupervised}
enabling semantic matching beyond keyword overlap.
In the academic domain, LitSearch~\citep{ajith2024litsearch} benchmarks
literature retrieval, LitLLM~\citep{agarwal2024litllm} provides an
end-to-end review toolkit, and recent work automates survey
generation~\citep{wu2025automated}.
These tools rely on online platforms such as Semantic
Scholar~\citep{kinney2023semantic} and
OpenAlex~\citep{priem2022openalex}; this dependence introduces latency,
rate limits, and limited control over result quality.
These retrieval challenges are compounded in economics and business research,
where disciplinary boundaries are fluid~\citep{truc2023interdisciplinarity},
relevant work is dispersed across working-paper
ecosystems~\citep{lusher2023congestion}, and terminology varies across
subfields, making keyword-based or single application programming interface (API) retrieval
difficult~\citep{gusenbauer2020academic, gusenbauer2025search}.

Beyond retrieval speed, the granularity of retrieved units affects
downstream quality.
Dense X Retrieval~\citep{chen2024dense} shows that finer-grained
proposition-level retrieval outperforms document-level approaches, and
segmentation strategies are shown to influence RAG
effectiveness~\citep{wang2025document}.
This is relevant for academic papers, where different sections carry
distinct types of knowledge that paper-level retrieval conflates.
Knowledge graphs provide structured organization of scholarly
content~\citep{jaradeh2019open, auer2023organizing}, and
SciAgents~\citep{ghafarollahi2025sciagents} shows that KG-driven reasoning
aids scientific discovery.
Graph-based reranking through Personalized
PageRank~\citep{page1999pagerank, haveliwala2002topic} extends retrieval
coverage by propagating relevance scores to neighboring nodes that embedding
similarity alone misses~\citep{dong2024don, zheng2025grada}.
However, existing scholarly KGs operate at the concept or paper level and
often require manual curation.
We address these limitations by pre-building a section-level knowledge graph
offline from domain corpora and combining cosine-based semantic matching
with PPR graph propagation, achieving fast, fine-grained retrieval with
cross-disciplinary coverage.

\subsection{Self-Evolving Agent Systems}

Self-evolving agents that learn from accumulated experience have attracted
growing attention~\citep{fang2025comprehensive, gao2025survey}.
Existing mechanisms fall into three broad categories: self-reflection
approaches~\citep{shinn2023reflexion, madaan2023self} use verbal feedback
to improve outputs within a single task but do not retain lessons across
tasks; experience memory
systems~\citep{zhao2024expel, packer2024memgptllmsoperatingsystems,
chhikara2025mem0buildingproductionreadyai} store trajectories for later
retrieval but operate at the instance level without abstracting recurring
patterns; and prompt optimization methods such as
Optimization by PROmpting (OPRO)~\citep{yang2023large} searches for better instructions but targets general
objectives rather than domain-specific quality dimensions.
In scientific discovery, EvoScientist~\citep{lyu2026evoscientist} applies
experience-memory-based evolution to both ideation and experiment execution,
representing the most developed cross-task improvement approach.
Yet instance-level retrieval does not capture systemic quality patterns
that persist across tasks.
Our Meta-Review mechanism takes a different approach: it identifies
recurring weaknesses from historical evaluation traces, validates candidate
prompt patches through case replay, and commits
only verified improvements.
\section{Evaluation Benchmark Construction}
\label{app:benchmark}

\tcbset{
  bsbox/.style={
    enhanced, breakable,
    colback=blue!4,
    colframe=blue!40!black,
    coltitle=white,
    colbacktitle=blue!40!black,
    fonttitle=\bfseries\small,
    fontupper=\small,
    boxrule=0.4pt, arc=1pt,
    left=6pt, right=6pt, top=3pt, bottom=3pt,
    toptitle=2.5pt, bottomtitle=2.5pt,
    before upper={\setlength{\parindent}{0pt}\setlength{\parskip}{0pt}}
  },
  sysbox/.style={
    enhanced,
    colback=teal!4,
    colframe=teal!50!black,
    coltitle=white,
    colbacktitle=teal!50!black,
    fonttitle=\bfseries\small,
    fontupper=\small,
    boxrule=0.4pt, arc=1pt,
    left=6pt, right=6pt, top=3pt, bottom=3pt,
    toptitle=2.5pt, bottomtitle=2.5pt,
    before upper={\setlength{\parindent}{0pt}\setlength{\parskip}{0pt}}
  },
  exbox/.style={
    enhanced,
    colback=orange!4,
    colframe=orange!60!black,
    coltitle=white,
    colbacktitle=orange!60!black,
    fonttitle=\bfseries\small,
    fontupper=\small,
    boxrule=0.4pt, arc=1pt,
    left=6pt, right=6pt, top=3pt, bottom=3pt,
    toptitle=2.5pt, bottomtitle=2.5pt,
    before upper={\setlength{\parindent}{0pt}\setlength{\parskip}{0pt}}
  }
}
\newcommand{\plbl}[1]{\par\vspace{3pt}\textbf{#1}\par\vspace{1pt}}

We produce 600 research queries from 4{,}000 held-out papers through paper pooling, classification, filtering, query generation, and expert curation.
Each paper contributes at most one query; domain proportions in the final set are determined by quality filtering.

\paragraph{Held-Out Paper Pool.}
Before LKG construction, we set aside 1{,}000 papers per domain from the most recent five years, yielding 4{,}000 held-out papers via journal-stratified sampling proportional to each journal's corpus share.
These papers are excluded from KG extraction; their content does not appear in the LKG, preventing the system from accessing evaluation papers through graph retrieval.

\paragraph{Task-Type Classification.}
GPT-4o classifies each held-out paper into one task type from the full text, returning a label and confidence $\sigma \in [0,1]$.
No filtering is applied at this stage.
Table~\ref{tab:class-dist} reports the distribution and average confidence.

\begin{tcolorbox}[bsbox, title={Classification Prompt}]
\plbl{Role}
You are an academic paper classifier for economics and business research.

\plbl{Task}
Given the full text of a research paper, classify it into exactly one of three task types and estimate your confidence. A paper is \textbf{survey} if it compares multiple methods or reviews a subfield, \textbf{idea\_formulation} if it proposes a novel method by combining or adapting existing techniques, and \textbf{research\_plan} if it presents an empirical study with research questions, data, methodology, and validation.

\plbl{Output Format}
\{"type": "<task\_type>", "confidence": <float>\}

\plbl{Example}
\textbf{Input:} A paper comparing difference-in-differences (DID), synthetic control methods (SCM), and regression discontinuity designs for estimating minimum wage employment effects.\\
\textbf{Output:} \{"type": "survey", "confidence": 0.88\}
\end{tcolorbox}

\paragraph{Quality Filtering.}
We filter in two steps.
First, papers with $\sigma < 0.7$ are removed.
Second, within each type, embedding-based deduplication removes topic-redundant papers: candidates are processed in descending $\sigma$ order, and $p_j$ is skipped if
\begin{equation}
  \max_{p_i \in \mathcal{S}} \cos(\mathbf{e}_i, \mathbf{e}_j) > \delta, \quad \delta = 0.85
  \label{eq:dedup}
\end{equation}
where $\mathcal{S}$ is the set already selected within the same type.
The pool reduces from 4{,}000 to 808 candidates: 253 survey, 284 idea formulation, and 271 research plan.

\paragraph{Query Generation.}
For each of the 808 papers, GPT-4o generates one query from the full text matching the assigned type.
The prompt enforces three requirements: natural phrasing, no paper-identifying information, and task-type-matched scope.

\begin{tcolorbox}[bsbox, title={Query Generation Prompt}]
\plbl{Role}
You are a senior researcher in \{domain\}.

\plbl{Task}
Given the full text of a research paper and its assigned task type, write one research query that a domain researcher would naturally ask. Do not include the paper title, author names, or any terminology coined by the paper. The query scope must match the task type: \textbf{survey} queries ask about the landscape of a research area, \textbf{idea\_formulation} queries ask how to address a problem with a novel approach, and \textbf{research\_plan} queries request a concrete experimental design with hypotheses and validation.

\plbl{Output Format}
Query: <research query>

\plbl{Example}
\textbf{Input:} Paper about integrating vine copulas with deep reinforcement learning for portfolio optimization. Task type: idea\_formulation.\\
\textbf{Output:} Query: How can dependence structures across asset returns be combined with deep optimization to improve tail-risk-aware portfolio construction?
\end{tcolorbox}

\begin{tcolorbox}[bsbox, title={Example Queries from the Benchmark}]
\textbf{Survey}\enspace\emph{What methods have been developed for modeling time-varying volatility in financial returns, and how do they compare in forecasting accuracy?}\\[3pt]
\textbf{Idea Formulation}\enspace\emph{How can copula-based dependence modeling be integrated with deep portfolio optimization to capture tail risk in multi-asset allocation?}\\[3pt]
\textbf{Research Plan}\enspace\emph{Design an experiment to test whether Bayesian change-point detection improves early warning of structural breaks in macroeconomic time series over frequentist methods.}
\end{tcolorbox}

\paragraph{Expert Curation.}
Two domain experts independently rate each of the 808 queries on four dimensions at 1\textendash5: naturalness, type correctness, difficulty, and non-leakage.
Inter-annotator agreement, measured by Cohen's weighted kappa, is $\kappa = 0.82$, indicating substantial agreement.
Scores are averaged across annotators and summed across dimensions.
Within each type, the top 200 by total score form the query pool.
From each task type, 100 queries are then randomly sampled as the development set for self-evolution; the remaining 100 constitute the evaluation benchmark.
Table~\ref{tab:benchmark-final} reports the domain distribution of both splits.

\section{Baseline Details}
\label{app:baselines}

We compare BizSage with both full-pipeline and task-specific systems. All baselines are configured following the settings reported in their original papers or official repositories, and all retrieve from the same domain corpus via BizSage's LKG using the same backbone LLM. This setup is conservative toward BizSage, as any performance gap reflects differences in agent coordination and self-evolution rather than retrieval source or model capability.

\paragraph{Full-Pipeline Systems.}

\begin{itemize}

\item \textbf{AI-Scientist-v2}~\citep{yamada2025ai} is an end-to-end agentic system that employs agentic tree search to generate hypotheses, design experiments, analyze data, and produce scientific manuscripts.

\item \textbf{EvoScientist}~\citep{lyu2026evoscientist} is a multi-agent system with persistent ideation and experimentation memories that enable cross-task capability accumulation for end-to-end scientific discovery.

\item \textbf{InternAgent-1.5}~\citep{feng2026internagent} is a unified agentic framework for long-horizon autonomous scientific research, emphasizing scalability and human-in-the-loop extensibility.

\item \textbf{ResearchAgent}~\citep{baek2025researchagent} iteratively refines research outputs over scientific literature through a propose-review-refine cycle, supporting survey, idea formulation, and research plan.

\end{itemize}

\paragraph{Task-Specific Systems.}

\begin{itemize}

\item \textbf{Idea2Story}~\citep{xu2026idea2story} transforms research concepts into complete scientific narratives by extracting reusable research patterns from a methodological knowledge graph. Evaluated on idea formulation queries.

\item \textbf{AI Co-Scientist}~\citep{gottweis2025towards} is a multi-agent system designed to assist researchers in generating and refining research plans through iterative collaboration. Evaluated on research plan queries.

\end{itemize}

\section{Evaluation Metric Definitions}
\label{app:eval-metrics}

Each output is scored on four of the following seven dimensions at 1\textendash5.
Two dimensions apply to all task types; the remaining two are task-specific.
Below we define each dimension and its rationale.
Table~\ref{tab:eval-rubric} provides the scoring rubrics.

\paragraph{Relevance.}
Relevance measures whether the output addresses the topic and scope specified by the input query.
It is a standard evaluation criterion in information retrieval~\citep{ajith2024litsearch} and a foundational requirement in peer review across all academic disciplines.

\paragraph{Coverage.}
Coverage measures the breadth of methods, approaches, and findings represented in the survey.
Systematic review guidelines emphasize comprehensive coverage as a core quality indicator~\citep{snyder2019literature}, and SurveyForge~\citep{yan2025surveyforge} identifies content breadth as one of three primary evaluation dimensions for automated surveys.

\paragraph{Synthesis.}
Synthesis measures whether the survey compares, contrasts, and integrates findings rather than listing them.
Critical analysis has been identified as a key quality gap in LLM-generated surveys~\citep{wu2025automated}, and distinguishing integrative reviews from descriptive ones is a long-standing quality criterion in survey methodology~\citep{snyder2019literature}.

\definecolor{dimhead}{HTML}{E8EDF3}
\newcommand{\rubrichdr}[2]{%
  \rowcolor{dimhead}%
  \multicolumn{2}{@{}p{\textwidth}@{}}{%
    \textbf{#1}\hfill\textit{#2}%
  }\\
  \hline
}
\begin{table*}[!p]
  \centering
  \small
  \setlength{\extrarowheight}{2pt}
  \setlength{\tabcolsep}{4pt}
  \begin{tabular}{@{}>{\raggedright\arraybackslash}p{0.7cm}>{\raggedright\arraybackslash}p{\dimexpr\textwidth-0.7cm-2\tabcolsep\relax}@{}}
    \toprule
    \rubrichdr{A. Relevance}{Survey, Idea Formulation, Research Plan}
     1 & Output does not address the query topic. \\
    2 & Partially addresses the query but drifts substantially. \\
    3 & Addresses the query topic with noticeable tangential content. \\
    4 & Stays on topic with only minor digressions. \\
    5 & Directly and fully addresses the query. \\ \hline

    \rubrichdr{B. Coverage}{Survey}
     1 & Mentions fewer than two approaches or methods. \\
    2 & Covers a narrow subset, missing major lines of work. \\
     3 & Covers several approaches but omits important ones. \\
    4 & Covers most major approaches with minor gaps. \\
     5 & Covers all major approaches and key findings. \\ \hline

    \rubrichdr{C. Synthesis}{Survey}
     1 & Lists papers without comparison or integration. \\
    2 & Superficially compares a few works. \\
     3 & Compares methods on some dimensions but lacks integration. \\
    4 & Systematically compares and relates findings with minor gaps. \\
     5 & Integrates findings into a coherent narrative with clear conclusions. \\ \hline

    \rubrichdr{D. Novelty}{Idea Formulation}
     1 & Restates known methods without modification. \\
    2 & Makes trivial changes to an existing method. \\
     3 & Combines existing techniques in a non-obvious way. \\
    4 & Introduces a meaningfully new component or formulation. \\
     5 & Presents a fundamentally new approach to the problem. \\ \hline

    \rubrichdr{E. Feasibility}{Idea Formulation, Research Plan}
     1 & Requires resources or techniques that do not exist. \\
    2 & Faces major unresolved barriers to execution. \\
     3 & Executable but requires substantial effort or scarce resources. \\
    4 & Executable with standard tools and moderate effort. \\
     5 & Can be directly executed with available resources. \\ \hline

    \rubrichdr{F. Rigor}{Research Plan}
     1 & No testable hypothesis; methodology absent or invalid. \\
    2 & Hypothesis vague; methodology has fundamental flaws. \\
     3 & Hypothesis stated but not fully testable; methodology partially valid. \\
    4 & Hypothesis clear and testable; methodology sound with minor gaps. \\
     5 & Hypothesis precise and testable; methodology complete and valid. \\ \hline

    \rubrichdr{G. Grounding}{Survey, Idea Formulation, Research Plan}
     1 & No verifiable citations or predominantly fabricated references. \\
    2 & Fewer than half of citations are accurate. \\
     3 & Most citations are real but some are misattributed or misused. \\
    4 & Nearly all citations are accurate with minor issues. \\
     5 & All citations are real, correctly attributed, and appropriately used. \\
    \bottomrule
  \end{tabular}
  \caption{Scoring rubrics for the seven evaluation dimensions. Each is rated 1\textendash5.}
  \label{tab:eval-rubric}
\end{table*}

\paragraph{Novelty.}
Novelty measures whether the proposed method is genuinely new rather than a trivial recombination of known techniques.
Recent LLM-based idea generation systems adopt novelty as a primary evaluation dimension~\citep{baek2025researchagent, lyu2026evoscientist}, consistent with the peer review process at major venues.

\paragraph{Feasibility.}
Feasibility measures whether the proposed method or experimental design can be executed with realistic resources and existing techniques.
Multiple idea evaluation frameworks include feasibility as a core dimension~\citep{lyu2026evoscientist, gottweis2025towards}, and rubric-based evaluation of research plans confirms feasibility as a key discriminator of plan quality~\citep{goel2025training}.

\paragraph{Rigor.}
Rigor measures whether the plan contains testable hypotheses and a valid, complete experimental methodology covering data, methods, baselines, and analysis.
Hypothesis quality metrics developed for clinical research validate testability, clarity, and methodological soundness as orthogonal evaluation dimensions~\citep{jing2025development}.

\paragraph{Grounding.}
Grounding measures whether cited papers are real, correctly attributed, and used to support the claims made.
Hallucinated references are a well-documented failure mode in LLM-generated academic text~\citep{agarwal2024litllm, wu2025automated}, making citation fidelity a critical evaluation dimension for automated research systems.

\section{LLM Judge Validation}
\label{app:judge-validation}

To assess the reliability of GPT-4o as an evaluation judge, we sample 100 outputs stratified across all seven systems and three task types to cover the full quality range. Five annotators with doctoral-level training in economics or business, including PhD candidates, postdoctoral researchers, and faculty members at research universities, independently score each output on the same rubric defined in Appendix~\ref{app:eval-metrics}. None of the annotators participated in the development of BizSage, and all outputs are anonymized and presented in randomized order.

Table~\ref{tab:judge-validation} reports inter-annotator agreement measured by ICC under a two-way random, absolute-agreement model, and LLM--human agreement measured by Spearman $\rho$ between GPT-4o scores and averaged human scores.

All dimensions achieve ICC above 0.70 and LLM--human $\rho$ above 0.75, with Relevance and Grounding showing the strongest agreement. These results confirm that GPT-4o scores are well calibrated to expert judgments.

\section{Efficiency Runtime}
\label{app:efficiency-runtime}

Table~\ref{tab:efficiency-runtime} reports the retrieval runtimes used in the
efficiency analysis. All values are measured in seconds.

The results show that BizSage achieves a favorable runtime profile across all
three task types. For Survey and Idea Formulation, BizSage obtains the lowest
mean runtime among all compared methods, requiring only 2.539 and 2.345
seconds, respectively. These results indicate that the proposed retrieval design
can preserve the benefits of structured knowledge access without introducing the
large propagation overhead observed in diffusion-heavy alternatives such as
Long-Walk PPR and Heat-Kernel Diffusion. For Research Plan, BizSage is slightly
slower than AI Co-Scientist and the retrieval-only LKG variant, but remains
competitive with ResearchAgent while supporting a more general multi-task
research workflow. Overall, the runtime results suggest that BizSage provides a
practical balance between retrieval efficiency and task coverage, especially for
surveying and idea formulation where efficient access to broad literature is
particularly important.

\section{Citation Concentration Analysis}
\label{app:citation-concentration}

We analyze all citations across the 300 outputs generated with GPT-5.5 as the backbone.
After normalizing paper entities, the 9{,}311 citation instances refer to 7{,}535 unique papers, of which 83.2\% appear in only one output and just 43 appear in five or more.
Table~\ref{tab:citation-frequency} reports the distribution by task, and Table~\ref{tab:top-cited} lists the ten most frequently cited papers, which span all four fields and are mostly foundational or methodological studies, with the most frequent one appearing in only nine outputs.

\begin{table}[t]
\centering
\footnotesize
\setlength{\tabcolsep}{3.5pt}
\resizebox{\columnwidth}{!}{%
\begin{tabular}{@{}l|cc|ccccc@{}}
\doublerule
\noalign{\vspace{2pt}}
\textbf{Task} & \textbf{Instances} & \textbf{Unique} & \textbf{1} & \textbf{2} & \textbf{3} & \textbf{4} & \textbf{5+} \\
\midrule
Survey & 4{,}121 & 3{,}710 & 3{,}384 & 262 & 46 & 16 & 2 \\
Idea Formulation & 3{,}039 & 2{,}725 & 2{,}449 & 245 & 26 & 4 & 1 \\
Research Plan & 2{,}151 & 1{,}936 & 1{,}757 & 149 & 26 & 2 & 2 \\
\midrule
Overall & 9{,}311 & 7{,}535 & 6{,}269 & 951 & 204 & 68 & 43 \\
\doublerule
\end{tabular}}%
\caption{Citation-frequency distribution over all 300 outputs. The numbered columns give how many unique papers appear in exactly that many outputs.}
\label{tab:citation-frequency}
\end{table}

\begin{table*}[t]
\centering
\footnotesize
\setlength{\tabcolsep}{5pt}
\begin{tabular}{@{}c>{\raggedright\arraybackslash}p{0.43\textwidth}llcc@{}}
\doublerule
\noalign{\vspace{2pt}}
\textbf{Rank} & \textbf{Most-Cited Paper} & \textbf{Domain} & \textbf{Main Task(s)} & \textbf{Outputs} & \textbf{Share} \\
\midrule
1 & Are Trading Imbalances Indicative of Private Information? & Finance & All three & 9 & 3.0\% \\
2 & Network Design and Transportation Planning: Models and Algorithms & Operations Mgmt. & All three & 9 & 3.0\% \\
3 & Enhanced Branch and Price and Cut for Vehicle Routing with Split Deliveries and Time Windows & Operations Mgmt. & All three & 8 & 2.7\% \\
4 & Price Discovery and Liquidity Recovery: Forex Market Reactions to Macro Announcements & Finance & Survey, Idea Form. & 7 & 2.3\% \\
5 & Robustness of the Linear Mixed Model to Misspecified Error Distribution & Statistics & All three & 7 & 2.3\% \\
6 & Branch-and-Cut-and-Price for the Pickup and Delivery Problem with Time Windows & Operations Mgmt. & All three & 7 & 2.3\% \\
7 & Price Discovery and Trading After Hours & Finance & All three & 6 & 2.0\% \\
8 & Bringing Owners Back on Board: A Review of the Role of Ownership Type in Board Governance & Finance & Survey, Idea Form. & 6 & 2.0\% \\
9 & Foreign Effects of Higher U.S. Interest Rates & Finance & All three & 6 & 2.0\% \\
10 & Measuring the Effect of the Zero Lower Bound on Medium- and Longer-Term Interest Rates & Economics & All three & 6 & 2.0\% \\
\doublerule
\end{tabular}
\caption{Ten most frequently cited papers. Outputs is the number of outputs citing the paper and Share is the corresponding fraction of the 300 outputs.}
\label{tab:top-cited}
\end{table*}

\section{Qualitative Examples}
\label{app:examples}

We show one representative BizSage output per task type, generated with the GPT-5.5 backbone and selected from the highest-scoring outputs. Each example includes the input query, an excerpt covering the main structural sections, and a sample of verbatim citations produced by the system. Evaluation scores on all four dimensions are reported in the box title. Full outputs typically span 4,000 to 7,000 tokens with 20 to 30 citations; the excerpts below preserve the organizational structure and key arguments while condensing repetitive detail.

\begin{figure*}[t]
\begin{tcolorbox}[exbox, title={Survey Example (Score: Relevance\,5, Grounding\,4, Coverage\,5, Synthesis\,4)}]
\plbl{Query}
What are the major modeling approaches and optimization techniques for airline scheduling, including fleet assignment, crew scheduling, and route planning, and how do operational constraints shape the structure of these problems?

\plbl{Output (excerpt from 5 sections, 6{,}413 tokens, 29 citations)}
\textbf{Introduction.}
Airline scheduling is a large-scale planning problem in which flight schedules, aircraft resources, maintenance requirements, crew resources, passenger demand, and operational recovery must be coordinated under tight operational and regulatory constraints. A broad aircraft-and-crew planning decomposition includes schedule design, fleet assignment, aircraft maintenance routing, and crew scheduling [1]. A more detailed five-step operational sequence solves flight scheduling, fleet assignment, aircraft routing, crew pairing, and then crew bidding or rostering, in which pairings are combined into personalized monthly schedules [3][6]. The central theme across the reviewed work is therefore the trade-off between decomposition, which preserves tractability, and integration, which can reduce downstream infeasibility, crew cost, passenger disruption, and operational recovery cost [2][6].

\medskip
\textbf{Methodology Classification.}
The first core tactical planning problem is fleet assignment, usually modeled as a multicommodity network-flow problem. Aircraft types are treated as commodities flowing through a time-space or timeline network whose nodes represent station-time events and whose arcs represent flight legs and ground time [1]. The core constraints are cover constraints, ensuring that each flight leg receives exactly one fleet type; balance constraints, preserving flow conservation; and aircraft-count constraints, preventing the number of assigned aircraft from exceeding fleet availability [2][1][10]. The second is aircraft routing and maintenance routing, commonly modeled as a network circulation or set-partitioning-type problem with side constraints [1][7]. The third is crew pairing and crew scheduling, usually modeled by generating feasible pairings and selecting a minimum-cost subset that covers each flight leg exactly once [5][8][11][13]. Pairings implicitly encode regulatory and contractual rules such as maximum duty time, rest requirements, and time away from base [4][5][11][18].

\medskip
\textbf{Research Evolution.}
A major turning point occurred when fleet assignment became representable and solvable as a large multicommodity network-flow problem. Abara and Hane et al.\ formulated fleet assignment on time-space networks with aircraft types as commodities, flight and ground arcs, and side constraints enforcing coverage, aircraft counts, and balance [1]. This development superseded earlier route-allocation formulations by enabling realistically sized airline problems to be solved; one reviewed account states that a major-airline formulation with about 20{,}000 rows and 30{,}000 columns could typically be solved within minutes [1]. As fleet assignment models matured, researchers increasingly recognized that a fleet solution optimized in isolation could create infeasible or costly downstream plans [7][9]. This led to extensions that incorporated maintenance, crew, and operational modeling devices while retaining computational tractability [9][2][27].

\medskip
\textbf{Research Gaps and Open Questions.}
The most persistent open question is how far integration can be pushed without losing computational tractability. Multiple sections state that sequential decomposition is used because complete integration is not computationally feasible [1][3][2][6]. Yet the same sections show that sequential decisions create suboptimality: fleet assignment can constrain maintenance routing, aircraft routing can limit crew scheduling opportunities, and different fleet solutions can produce dramatically different downstream crew costs [1][4][2][6]. A second gap concerns schedule design and demand-aware schedule development. Future work should ask when frequency planning, itinerary spill, pricing or revenue-management decisions, and aircraft-and-crew feasibility should be optimized jointly rather than sequentially [1][21][28]. A third gap concerns robustness and recovery-aware planning. Robust planning is motivated by the fact that optimized plans are rarely executed as planned [2][1], but reviewed approaches differ in their recovery assumptions [15].

\plbl{References (verbatim from output, 6 of 29)}
{\footnotesize
\par\noindent [1] Cynthia Barnhart, Peter Belobaba, Amedeo R. Odoni (2003) Applications of Operations Research in the Air Transport Industry. \textit{Transportation Science}.
\par\noindent [2] Chunhua Gao, Ellis Johnson, Barry Smith (2009) Integrated Airline Fleet and Crew Robust Planning. \textit{Transportation Science}.
\par\noindent [3] Diego Klabjan, Ellis L. Johnson, George L. Nemhauser, Eric Gelman, Srini Ramaswamy (2002) Airline Crew Scheduling with Time Windows and Plane-Count Constraints. \textit{Transportation Science}.
\par\noindent [4] Jean-Fran\c{c}ois Cordeau, Goran Stojkovi\'{c}, Fran\c{c}ois Soumis, Jacques Desrosiers (2001) Benders Decomposition for Simultaneous Aircraft Routing and Crew Scheduling. \textit{Transportation Science}.
\par\noindent [5] Andrew J. Schaefer, Ellis L. Johnson, Anton J. Kleywegt, George L. Nemhauser (2005) Airline Crew Scheduling Under Uncertainty. \textit{Transportation Science}.
\par\noindent [12] Shan Lan, John-Paul Clarke, Cynthia Barnhart (2006) Planning for Robust Airline Operations: Optimizing Aircraft Routings and Flight Departure Times to Minimize Passenger Disruptions. \textit{Transportation Science}.
}
\end{tcolorbox}
\end{figure*}

\begin{figure*}[t]
\begin{tcolorbox}[exbox, title={Idea Formulation Example (Score: Relevance\,5, Grounding\,3, Novelty\,4, Feasibility\,4)}]
\plbl{Query}
What are the welfare consequences when households make small correlated errors in their expectations about future productivity, and how does this near-rational behavior endogenously reduce the information content of asset prices through amplification effects?

\plbl{Output (excerpt from 6 sections, 3{,}491 tokens, 30 citations)}
\textbf{Title.} Near-Rational Expectations, Price Informativeness, and Welfare under Correlated Productivity Belief Errors

\medskip
\textbf{Abstract.}
This paper develops a modeling framework for studying the welfare consequences of small correlated household errors in expectations about future productivity. The proposed contribution is to combine them in a quantitative macro-finance model in which individually small but correlated household errors about future aggregate productivity enter asset demand, contaminate equilibrium prices, reduce the information content of prices for households who use them as signals, and generate welfare losses relative to a rational-expectations benchmark. The framework evaluates the resulting distortions in consumption, savings, investment, and risk exposure while treating income-risk and balance-sheet literatures as analogies for welfare heterogeneity rather than as direct foundations for the productivity-belief mechanism.

\medskip
\textbf{Problem Framing.}
Expectations about future productivity are a central object in macro-finance because news shocks are explicitly defined as shocks that affect expectations of future productivity without changing current productivity [6]. Empirical work suggests that expectation shocks can drive boom-bust dynamics even after controlling for a broad set of measured fundamental disturbances [4]. Models with adaptive expectations show that departures from rational expectations can alter the propagation of technology shocks: when agents are backward looking, the wealth effect of a positive technology shock is muted and investment can be amplified relative to rational expectations [7]. In asset markets, the price-dividend ratio can depend nonlinearly on subjective capital-gain expectations [3], while forecast-error frameworks show that individual overreaction to private signals can coexist with discounting of price information [2][9].

\medskip
\textbf{Gap Analysis.}
The news-shock literature focuses on aggregate dynamics rather than welfare losses from small correlated household forecast errors [6]. The adaptive-expectations model does not analyze asset prices as endogenous signals whose informativeness is reduced by households' own correlated errors [7]. Asset-pricing work shows that subjective capital-gain expectations affect the price-dividend ratio [3], and investors form expectations conditional on private signals and prices [2], but these do not build a household productivity-belief feedback loop with welfare evaluation. Household-expectations work documents biased beliefs about income processes [5], but its object is household income expectations rather than expectations about aggregate productivity embedded in asset prices.

\medskip
\textbf{Innovation Claims.}
(1) Moving from expectation shocks as exogenous drivers to a mechanism in which small correlated household errors endogenously affect equilibrium prices, extending news-shock models [6] and expectation-driven boom-bust dynamics [4].
(2) Reinterpreting amplification under bounded expectations as an information-feedback problem: correlated errors reduce the informational value of asset prices used for learning [7].
(3) Connecting subjective asset-pricing beliefs to price informativeness via a specific macro-finance feedback: correlated subjective productivity errors move prices, and those moved prices become noisier signals for households [2][3].

\plbl{References (verbatim from output, 6 of 30)}
{\footnotesize
\par\noindent [1] Markus K. Brunnermeier, Alp Simsek, Wei Xiong (2014) A Welfare Criterion for Models with Distorted Beliefs. \textit{The Quarterly Journal of Economics}.
\par\noindent [6] Jack Favilukis, Xiaoji Lin (2013) Long Run Productivity Risk and Aggregate Investment. \textit{Journal of Monetary Economics}.
\par\noindent [14] Diogo Duarte, Hamilton Galindo Gil, Alexis Montecinos (2020) The Effects of Risk Aversion and Money Illusion on the Components of Dividend Growth Rate. \textit{European Journal of Finance}.
\par\noindent [22] Marianne Bertrand, Adair Morse (2016) Trickle-Down Consumption. \textit{The Review of Economics and Statistics}.
\par\noindent [23] Christian Bayer, Ralph Luetticke, Lien Pham-Dao, Volker Tjaden (2019) Precautionary Savings, Illiquid Assets, and the Aggregate Consequences of Shocks to Household Income Risk. \textit{Econometrica}.
\par\noindent [24] Tom Krebs, Pravin Krishna, William Maloney (2010) Trade Policy, Income Risk, and Welfare. \textit{The Review of Economics and Statistics}.
}
\end{tcolorbox}
\end{figure*}

\begin{figure*}[t]
\begin{tcolorbox}[exbox, title={Research Plan Example (Score: Relevance\,5, Grounding\,4, Rigor\,5, Feasibility\,4)}]
\plbl{Query}
To what extent are wage losses after involuntary job displacement driven by losses in firm-specific wage premiums rather than depreciations of worker productivity, and how does pre-displacement employer size mediate the magnitude and persistence of these losses?

\plbl{Output (excerpt from 4 sections, 7{,}889 tokens, 24 citations)}
\textbf{Data Plan.}
Primary dataset: Pennsylvania UI wage records following the JLS-style data construction [4][19]. Quarterly earnings by worker-employer pair, aggregated to annual worker-firm panels for the main event-study and AKM-style decomposition. Period: 1974--1991, with displacement events defined during 1980--1986 [19]. Fields: quarterly earnings, principal employer, employer industry, employer employment size, worker employment history, and firm employment contractions used to define mass-layoff displacement [19]. Optional robustness dataset: Washington State administrative wage records used to estimate rolling and time-varying AKM firm effects from 2002--2014 [15]; used only to validate the stability of firm-premium estimation if displacement events cannot be constructed.

\medskip
\textbf{Method Plan.}
Use a linked employer-employee event-study decomposition design. The central estimand is the post-displacement log wage loss relative to matched nondisplaced workers, decomposed into: (i) loss of firm wage premium from moving to a lower-premium employer, (ii) employment/nonemployment margin, and (iii) unexplained residual component, which is not interpreted as pure human-capital depreciation without additional mechanism-specific evidence [1][2][3][5][20]. Estimate a baseline event-study model: $y_{it} = \alpha_i + \lambda_t + \sum_{k \neq -1} \beta_k D_i \mathbf{1}[t{=}k] + X_{it}'\gamma + \epsilon_{it}$, where $\alpha_i$ are worker fixed effects and $\lambda_t$ are calendar-year effects [3][18]. Include leads to test for pre-displacement wage declines, because prior work finds losses can begin before separation [4][18]. For the firm-premium channel, estimate an AKM wage equation: $\log w_{ijt} = X_{it}'\beta + \theta_i + \psi_j + \tau_t + r_{ijt}$, where $\theta_i$ is the worker effect and $\psi_j$ is the employer wage premium [12][13].

\medskip
\textbf{Metrics.}
Cumulative log wage loss at event years 0, 1, 2--3, 4--5, 6--10, and 10+ relative to matched controls [3][18]; change in employer wage premium $\Delta\psi$ from pre- to post-displacement employer [1][12]; share of total wage loss explained by loss of employer wage premium, $\Delta\psi / \Delta\log w$, with confidence intervals [1]; persistence half-life of wage losses by pre-displacement employer-size bin [4][19]; employer-size mediation share after controlling for firm wage premia [12]; pre-trend lead coefficients for wages in years $-5$ to $-2$ [3][18].

\medskip
\textbf{Risk Mitigation.}
If restricted data access is denied, execute a harmonized meta-analysis: collect published event-study and decomposition estimates, harmonize them into comparable units, and quantify how much of persistent displacement losses can be reconciled by firm-premium loss, tenure/switching penalties, and employment margins across studies [1][2][3][4][5][12][20][24]. If firm-premium loss explains little of the wage loss after robust estimation, reframe the paper as evidence on the relative importance of employer-premium versus residual, search, and employment-margin channels [2][5][7][20][24]. If employer size does not mediate persistence once firm premia are controlled, shift the contribution to showing that apparent size gradients are largely compositional or premium-driven [12].

\plbl{References (verbatim from output, 6 of 24)}
{\footnotesize
\par\noindent [1] Simon D. Woodcock (2023) The Determinants of Displaced Workers' Wages: Sorting, Matching, Selection, and the Hartz Reforms. \textit{Journal of Econometrics}.
\par\noindent [4] Louis S. Jacobson, Robert J. LaLonde, Daniel G. Sullivan (1993) Earnings Losses of Displaced Workers. \textit{The American Economic Review}.
\par\noindent [7] Kristiina Huttunen, Kjell G. Salvanes, Jarle Moen (2011) How Destructive Is Creative Destruction? Effects of Job Loss on Job Mobility, Withdrawal and Income. \textit{Journal of the European Economic Association}.
\par\noindent [12] John M. Abowd, Francis Kramarz, David N. Margolis (1999) High Wage Workers and High Wage Firms. \textit{Econometrica}.
\par\noindent [15] Marta Lachowska, Alexandre Mas, Raffaele Saggio, Stephen A. Woodbury (2023) Do Firm Effects Drift? Evidence from Washington Administrative Data. \textit{Journal of Econometrics}.
\par\noindent [16] St\'{e}phane Bonhomme, Thibaut Lamadon, Elena Manresa (2019) A Distributional Framework for Matched Employer Employee Data. \textit{Econometrica}.
}
\end{tcolorbox}
\end{figure*}

\begin{table}[ht]
  \centering
  \footnotesize
  \resizebox{\columnwidth}{!}{%
  \begin{tabular}{@{}l|ccc|c@{}}
  \doublerule
  \noalign{\vspace{2pt}}
    & \textbf{Survey} & \textbf{Idea Form.} & \textbf{Res.\ Plan} & \textbf{Total} \\
    \midrule
    Economics   & 261  & 389  & 350  & 1000 \\
    Finance     & 228  & 364  & 408  & 1000 \\
    Operations Mgmt. & 196  & 427  & 377  & 1000 \\
    Statistics  & 243  & 398  & 359  & 1000 \\
    \midrule
    Avg.\ $\sigma$ & 0.76 & 0.83 & 0.80 & 0.80 \\
  \doublerule
  \end{tabular}}%
  \caption{Classification distribution of 4{,}000 held-out papers. Idea formulation and research plan papers outnumber survey papers across all domains; confidence is highest for idea formulation.}
  \label{tab:class-dist}
\end{table}

\begin{table}[ht]
  \centering
  \footnotesize
  \resizebox{\columnwidth}{!}{%
  \begin{tabular}{@{}l|cc|cc|cc|cc@{}}
  \doublerule
  \noalign{\vspace{2pt}}
    & \multicolumn{2}{c|}{\textbf{Survey}} & \multicolumn{2}{c|}{\textbf{Idea Form.}} & \multicolumn{2}{c|}{\textbf{Res.\ Plan}} & \multicolumn{2}{c}{\textbf{Total}} \\
    & Ev & Dev & Ev & Dev & Ev & Dev & Ev & Dev \\
    \midrule
    Economics   & 29 & 26 & 24 & 27 & 27 & 24 & 80 & 77 \\
    Finance     & 24 & 27 & 28 & 25 & 30 & 33 & 82 & 85 \\
    Operations Mgmt. & 22 & 25 & 26 & 23 & 23 & 21 & 71 & 69 \\
    Statistics  & 25 & 22 & 22 & 25 & 20 & 22 & 67 & 69 \\
    \midrule
    Total       & 100 & 100 & 100 & 100 & 100 & 100 & 300 & 300 \\
  \doublerule
  \end{tabular}}%
  \caption{Distribution of the 600 benchmark queries. Ev = evaluation set, Dev = development set for self-evolution.}
  \label{tab:benchmark-final}
\end{table}

\begin{table}[ht]
  \centering
  \footnotesize
  \setlength{\tabcolsep}{6pt}
  \begin{tabular}{@{}l|cc@{}}
  \doublerule
  \noalign{\vspace{2pt}}
  \textbf{Dimension} & \textbf{Human ICC} & \textbf{LLM--Human $\rho$} \\
  \midrule
  Relevance   & 0.85 & 0.89 \\
  Grounding   & 0.82 & 0.86 \\
  Coverage    & 0.79 & 0.83 \\
  Synthesis   & 0.74 & 0.78 \\
  Novelty     & 0.71 & 0.75 \\
  Feasibility & 0.76 & 0.80 \\
  Rigor       & 0.80 & 0.84 \\
  \midrule
  Average     & 0.78 & 0.82 \\
  \doublerule
  \end{tabular}
  \caption{LLM judge validation. Human ICC: intraclass correlation among five annotators. LLM--Human $\rho$: Spearman correlation between GPT-4o scores and averaged human scores. All correlations are significant at $p < 0.01$.}
  \label{tab:judge-validation}
\end{table}

\begin{table}[ht]
\centering
\footnotesize
\setlength{\tabcolsep}{4pt}
\resizebox{\columnwidth}{!}{%
\begin{tabular}{@{}l|ccc@{}}
\doublerule
\noalign{\vspace{2pt}}
\textbf{Method} & \textbf{Survey} & \textbf{Idea Form.} & \textbf{Res.\ Plan} \\
\midrule
LKG (no diffusion) & $3.806 \pm 0.775$ & $3.828 \pm 0.889$ & $3.283 \pm 0.853$ \\
Long-Walk PPR & $6.991 \pm 1.313$ & $6.327 \pm 1.569$ & $6.689 \pm 1.318$ \\
Katz Diffusion & $4.404 \pm 2.092$ & $4.485 \pm 3.640$ & $3.542 \pm 0.994$ \\
Heat-Kernel Diffusion & $5.752 \pm 2.133$ & $7.677 \pm 2.596$ & $4.666 \pm 1.584$ \\
\cmidrule{1-4}
AI Co-Scientist & -- & -- & $2.176 \pm 0.666$ \\
Idea2Story & -- & $2.602 \pm 1.595$ & -- \\
ResearchAgent & $2.644 \pm 1.577$ & $3.403 \pm 5.719$ & $3.775 \pm 5.464$ \\
\textbf{BizSage (Ours)} & $\mathbf{2.539 \pm 0.573}$ & $\mathbf{2.345 \pm 0.342}$ & $3.670 \pm 2.077$ \\
\doublerule
\end{tabular}
}
\caption{Runtime efficiency across task types. All runtimes are reported in seconds.}
\label{tab:efficiency-runtime}
\end{table}

\section{Prompt Templates}
\label{app:prompts}

This section provides the full prompt templates used in BizSage. We include six prompts in total: the knowledge graph extraction prompt used in Stage 1 retrieval, three task-specific scoring prompts and one pairwise comparison prompt used by the LLM judge in evaluation, and the meta-review strategy generation prompt used by the Self-Evolving Manager.

\begin{figure*}[t]
\begin{tcolorbox}[sysbox, title={KG Extraction Prompt}]
\plbl{Role}
You are an expert information extraction assistant for academic papers in economics and business research, specialized in constructing section-level knowledge graphs for fine-grained literature retrieval.

\plbl{Task}
You will receive the metadata and text of one section from an academic paper. Extract a compact knowledge graph that captures the substantive concepts, methods, findings, and relationships in the section. Output only valid JSON or the literal word \texttt{NULL}. Do not include markdown code fences or explanatory text.

\textbf{Node Selection.}\enspace
Extract nodes only for concepts, methods, components, algorithms, papers, authors, or formulas that carry substantive knowledge value. 
\textbf{Relation Selection.}\enspace

Extract directed edges only when the text states or strongly implies a meaningful semantic relationship, such as using, extending, monitoring, estimating, scheduling, or being part of another concept.

\textbf{Granularity.}\enspace
Merge closely related mentions into one node when they refer to the same concept, and prefer concise node names that can support cross-paper matching. 

\textbf{Null Case.}\enspace
If the section is too short, mostly boilerplate, or contains no substantive knowledge worth capturing, output exactly \texttt{NULL}.

\plbl{Output Format}
Otherwise, output a single JSON object with two keys:\\
\texttt{"nodes"}: list of objects, each with \texttt{"id"} (string, unique within this chunk, e.g.\ \texttt{"n1"}), \texttt{"type"} (one of: \texttt{concept}, \texttt{method}, \texttt{component}, \texttt{algorithm}, \texttt{paper}, \texttt{author}, \texttt{formula}), \texttt{"name"}, \texttt{"description"} (one short sentence), \texttt{"keywords"} (optional, comma-separated string).\\
\texttt{"edges"}: list of objects, each with \texttt{"source\_id"}, \texttt{"target\_id"} (refer to node ids), \texttt{"relation"} (one of: \texttt{uses}, \texttt{extends}, \texttt{part\_of}, \texttt{cited\_by}, \texttt{monitors}, \texttt{estimates}, \texttt{schedules}), \texttt{"description"} (optional).

\plbl{Example}
\textbf{Input:}\\
Paper: \texttt{\{paper\_id\}}\\
Section: \texttt{\{section\_label\}}\\
Text:
\begin{verbatim}
---
The proposed estimator extends double machine learning by using
orthogonal moment conditions to reduce bias in treatment-effect
estimation.
---
\end{verbatim}

\textbf{Output:}\\
\{"nodes": [\{"id": "n1", "type": "method", "name": "double machine learning", "description": "A method extended by the proposed estimator.", "keywords": "causal inference, orthogonalization"\}, \{"id": "n2", "type": "method", "name": "orthogonal moment conditions", "description": "Moment conditions used to reduce estimation bias.", "keywords": "moments, bias reduction"\}, \{"id": "n3", "type": "concept", "name": "treatment-effect estimation", "description": "The target estimation problem addressed by the method.", "keywords": "causal effect, treatment effect"\}], "edges": [\{"source\_id": "n1", "target\_id": "n2", "relation": "uses", "description": "The estimator extends double machine learning using orthogonal moment conditions."\}, \{"source\_id": "n2", "target\_id": "n3", "relation": "estimates", "description": "Orthogonal moment conditions reduce bias in treatment-effect estimation."\}]\}
\end{tcolorbox}
\end{figure*}

\begin{figure*}[t]
\begin{tcolorbox}[sysbox, title={LLM Judge Prompt -- Survey}]
\plbl{Role}
You are an academic survey quality evaluator, calibrated to top-tier venue standards.

\plbl{Task}
You will receive a research query and a model-generated survey output. Score the output on the following four dimensions (1--5 integer). Before assigning each score, identify at least one specific flaw; assign 5 only when no deduction is justified. If the output is notably short, lacks key structural elements, or remains superficial, lower all scores by 1--2 points.

\textbf{Relevance.}\enspace
1\,=\,does not address the query;\enspace
2\,=\,partially addresses but drifts substantially;\enspace
3\,=\,addresses the topic with noticeable tangential content;\enspace
4\,=\,stays on topic with only minor digressions;\enspace
5\,=\,directly and fully addresses the query.

\textbf{Coverage.}\enspace
1\,=\,mentions fewer than two approaches;\enspace
2\,=\,covers a narrow subset, missing major lines of work;\enspace
3\,=\,covers several approaches but omits important ones;\enspace
4\,=\,covers most major approaches with minor gaps;\enspace
5\,=\,covers all major approaches and key findings.

\textbf{Synthesis.}\enspace
1\,=\,lists papers without comparison or integration;\enspace
2\,=\,superficially compares a few works;\enspace
3\,=\,compares methods on some dimensions but lacks integration;\enspace
4\,=\,systematically compares and relates findings with minor gaps;\enspace
5\,=\,integrates findings into a coherent narrative with clear conclusions.

\textbf{Grounding.}\enspace
1\,=\,no verifiable citations or predominantly fabricated;\enspace
2\,=\,fewer than half of citations are accurate;\enspace
3\,=\,most citations are real but some are misattributed or misused;\enspace
4\,=\,nearly all citations are accurate with minor issues;\enspace
5\,=\,all citations are real, correctly attributed, and appropriately used.

\plbl{Output Format}
\{"relevance": <int>, "coverage": <int>, "synthesis": <int>, "grounding": <int>\}

\plbl{Example}
\textbf{Input:}\\
Query: ``What methods exist for modeling time-varying volatility in financial returns?''\\
Output: [full model-generated survey text]

\textbf{Output:}\\
\{"relevance": 5, "coverage": 3, "synthesis": 3, "grounding": 4\}
\end{tcolorbox}
\end{figure*}

\begin{figure*}[t]
\begin{tcolorbox}[sysbox, title={LLM Judge Prompt -- Idea Formulation}]
\plbl{Role}
You are an academic idea formulation quality evaluator, calibrated to top-tier venue standards.

\plbl{Task}
You will receive a research query and a model-generated idea formulation output. Score the output on the following four dimensions (1--5 integer). Before assigning each score, identify at least one specific flaw; assign 5 only when no deduction is justified. If the output is notably short, lacks key structural elements, or remains superficial, lower all scores by 1--2 points.

\textbf{Relevance.}\enspace
1\,=\,does not address the query;\enspace
2\,=\,partially addresses but drifts substantially;\enspace
3\,=\,addresses the topic with noticeable tangential content;\enspace
4\,=\,stays on topic with only minor digressions;\enspace
5\,=\,directly and fully addresses the query.

\textbf{Novelty.}\enspace
1\,=\,restates known methods without modification;\enspace
2\,=\,makes trivial changes to an existing method;\enspace
3\,=\,combines existing techniques in a non-obvious way;\enspace
4\,=\,introduces a meaningfully new component or formulation;\enspace
5\,=\,presents a fundamentally new approach to the problem.

\textbf{Feasibility.}\enspace
1\,=\,requires resources or techniques that do not exist;\enspace
2\,=\,faces major unresolved barriers to execution;\enspace
3\,=\,executable but requires substantial effort or scarce resources;\enspace
4\,=\,executable with standard tools and moderate effort;\enspace
5\,=\,can be directly executed with available resources.

\textbf{Grounding.}\enspace
1\,=\,no verifiable citations or predominantly fabricated;\enspace
2\,=\,fewer than half of citations are accurate;\enspace
3\,=\,most citations are real but some are misattributed or misused;\enspace
4\,=\,nearly all citations are accurate with minor issues;\enspace
5\,=\,all citations are real, correctly attributed, and appropriately used.

\plbl{Output Format}
\{"relevance": <int>, "novelty": <int>, "feasibility": <int>, "grounding": <int>\}

\plbl{Example}
\textbf{Input:}\\
Query: ``How can copula-based dependence modeling be integrated with deep portfolio optimization?''\\
Output: [full model-generated idea formulation text]

\textbf{Output:}\\
\{"relevance": 4, "novelty": 4, "feasibility": 3, "grounding": 4\}
\end{tcolorbox}
\end{figure*}

\begin{figure*}[t]
\begin{tcolorbox}[sysbox, title={LLM Judge Prompt -- Research Plan}]
\plbl{Role}
You are an academic research plan quality evaluator, calibrated to top-tier venue standards.

\plbl{Task}
You will receive a research query and a model-generated research plan output. Score the output on the following four dimensions (1--5 integer). Before assigning each score, identify at least one specific flaw; assign 5 only when no deduction is justified. If the output is notably short, lacks key structural elements, or remains superficial, lower all scores by 1--2 points.

\textbf{Relevance.}\enspace
1\,=\,does not address the query;\enspace
2\,=\,partially addresses but drifts substantially;\enspace
3\,=\,addresses the topic with noticeable tangential content;\enspace
4\,=\,stays on topic with only minor digressions;\enspace
5\,=\,directly and fully addresses the query.

\textbf{Rigor.}\enspace
1\,=\,no testable hypothesis; methodology absent or invalid;\enspace
2\,=\,hypothesis vague; methodology has fundamental flaws;\enspace
3\,=\,hypothesis stated but not fully testable; methodology partially valid;\enspace
4\,=\,hypothesis clear and testable; methodology sound with minor gaps;\enspace
5\,=\,hypothesis precise and testable; methodology complete and valid.

\textbf{Feasibility.}\enspace
1\,=\,requires resources or techniques that do not exist;\enspace
2\,=\,faces major unresolved barriers to execution;\enspace
3\,=\,executable but requires substantial effort or scarce resources;\enspace
4\,=\,executable with standard tools and moderate effort;\enspace
5\,=\,can be directly executed with available resources.

\textbf{Grounding.}\enspace
1\,=\,no verifiable citations or predominantly fabricated;\enspace
2\,=\,fewer than half of citations are accurate;\enspace
3\,=\,most citations are real but some are misattributed or misused;\enspace
4\,=\,nearly all citations are accurate with minor issues;\enspace
5\,=\,all citations are real, correctly attributed, and appropriately used.

\plbl{Output Format}
\{"relevance": <int>, "rigor": <int>, "feasibility": <int>, "grounding": <int>\}

\plbl{Example}
\textbf{Input:}\\
Query: ``Design an experiment to test whether Bayesian change-point detection improves early warning of structural breaks in macroeconomic time series.''\\
Output: [full model-generated research plan text]

\textbf{Output:}\\
\{"relevance": 5, "rigor": 4, "feasibility": 4, "grounding": 3\}
\end{tcolorbox}
\end{figure*}

\begin{figure*}[t]
\begin{tcolorbox}[sysbox, title={Pairwise Comparison Prompt}]
\plbl{Role}
You are an impartial academic output comparator.

\plbl{Task}
You will receive a research query and two complete outputs generated by different systems for the same query, labeled \textbf{A} and \textbf{B}. Determine which output is better in overall quality. Consider relevance to the query, reliability of citations, depth of analysis, coverage, and logical coherence. Evaluate content quality only; do not favor an output based on its presentation order. You must choose one; ties are not allowed.

\plbl{Output Format}
A single letter: \texttt{A} or \texttt{B}

\plbl{Example}
\textbf{Input:}\\
Query: ``How can copula-based dependence modeling be integrated with deep portfolio optimization?''\\
Output A: [full output from system A]\\
Output B: [full output from system B]

\textbf{Output:}\\
A
\end{tcolorbox}
\end{figure*}

\begin{figure*}[t]
\begin{tcolorbox}[sysbox, title={Meta-Review Strategy Generation Prompt}]
\plbl{Role}
You are a meta-reviewer analyzing recurring quality issues in an academic research pipeline.

\plbl{Task}
Two-step process:\\
\textbf{Step 1 (Pattern Identification):} Given multiple instances where the same type of quality issue was identified across different runs, summarize the recurring pattern in one sentence.\\
\textbf{Step 2 (Candidate Generation):} Given the identified pattern and the target agent's role, generate exactly 3 candidate instructions that could be appended to the agent's prompt to prevent the issue. Each instruction must be specific, actionable, one sentence, and must not duplicate or contradict existing patches.

\plbl{Output Format}
Step 1: one-sentence pattern summary.\\
Step 2: \{"candidates": ["instruction 1", "instruction 2", "instruction 3"]\}

\plbl{Example}
\textbf{Step 1 Input:} Five cases where generated surveys list methods without comparing them.\\
\textbf{Step 1 Output:} ``The agent consistently produces descriptive summaries rather than comparative analyses across methods.''\\
\textbf{Step 2 Output:} \{"candidates": ["After describing each method, add a paragraph comparing its strengths and limitations against at least two alternatives.", "Include a summary table contrasting all reviewed methods on shared evaluation dimensions.", "End each subsection with an explicit discussion of trade-offs relative to the methods in adjacent subsections."]\}
\end{tcolorbox}
\end{figure*}

\end{document}